\documentclass[letterpaper]{article} 
\usepackage{multirow}
\usepackage{pifont}
\usepackage{framed}
\usepackage{amsmath}
\usepackage{amssymb}
\usepackage[table]{xcolor}
\usepackage{aaai2027}  
\usepackage[hyphens]{url}  
\usepackage{graphicx} 
\usepackage{natbib}  
\usepackage{caption} 
\usepackage{algorithm}
\usepackage{algorithmic}

\usepackage{newfloat}
\usepackage{listings}
\DeclareCaptionStyle{ruled}{labelfont=normalfont,labelsep=colon,strut=off} 
\floatstyle{ruled}
\newfloat{listing}{tb}{lst}{}
\floatname{listing}{Listing}

\usepackage{booktabs}

\title{Multimodal In-Context Learning with Explicit Rule Induction}
\author{
    Yuquan Li\textsuperscript{\rm 1}, Xinyang Jiang\textsuperscript{\rm 2}, Yuhao Xue\textsuperscript{\rm 1}, Cairong Zhao\textsuperscript{\rm 1}
}
\affiliations{
    \textsuperscript{\rm 1}Tongji University\\

    \textsuperscript{\rm 2}Microsoft Research Asia\\
    proceedings-questions@aaai.org
}

\begin{document}

\maketitle

\begin{abstract}
    A central challenge in multimodal in‑context learning (MICL) arises from tasks that are ill‑defined without context and require uncovering the underlying rules from demonstrations. Existing methods predominantly treat MICL as direct input–output mapping, overlooking the importance of explicitly inducing and reasoning over such rules, limiting both performance and interpretability. 
    We propose ERIM, Explicit Rule-Induction for Multimodal In-Context Learning, a unified framework that enhances MICL by explicitly inducing the latent rules governing demonstration samples. Leveraging the reasoning ability of multi-modal large language model (MLLM), ERIM constructs rule-centric MICL datasets, introduces explicit supervision for rule induction, and fine-tunes  MLLM to perform explicit rule-induction via a two-stage training strategy that combines mixed cold-start supervised fine-tuning with reinforcement learning. 
    Extensive experiments show consistent performance improvements across diverse MICL settings and demonstrate that correctly inducing underlying rules directly leads to stronger and more stable performance. 
\end{abstract}


\section{Introduction}


In‑context learning (ICL)~\cite{NEURIPS2020_1457c0d6,wang2025context,dong2024survey} has emerged as a key capability of large language models, enabling training‑free task adaptation through a small number of contextual demonstrations. With the advent of multimodal models~\cite{chen2024internvl}, this paradigm has recently been extended to multimodal in‑context learning (MICL)~\cite{jiang2024many,chen2025can}. However, existing MICL evaluations largely focus on few‑shot variants of standard vision tasks, which have been shown to be weak proxies for genuine context‑dependent learning~\cite{chen2025true}. In such settings, performance gains may arise from parametric knowledge or superficial pattern imitation rather than true adaptation to the provided demonstrations~\cite{zong2025vl}.

\begin{figure}
    \centering
    \includegraphics[width=\linewidth]{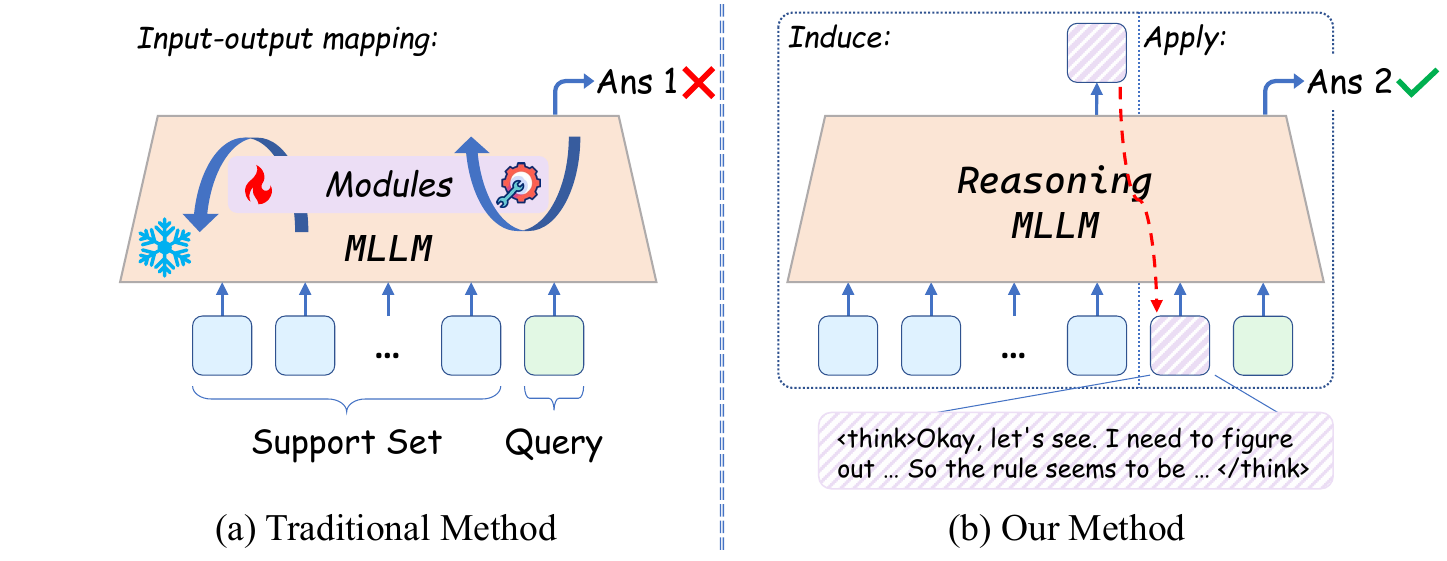}
    \caption{Comparison of MICL paradigms. (a) Existing methods treat MICL as a direct input-output mapping.
    (b) Our approach explicitly induces a rule from the support set and applies it to the query.}
    \label{fig:insight}
\end{figure}

To address this limitation, prior work distinguishes between two categories of MICL tasks. Case I consists of standard vision tasks that remain well‑defined even in the absence of demonstrations. In contrast, Case II tasks are governed by an underlying rule that is shared across examples and cannot be inferred without contextual information~\cite{huang2025mimicking}. Consequently, demonstrations in Case II jointly define how the task should be solved, requiring models to extract, abstract, and generalize task‑defining rules from the context and apply the induced rule to unseen queries. As a result, Case II more faithfully reflects the core notion of in‑context learning and provides a more stringent test of whether multimodal models truly learn from context rather than rely on default inference behaviors~\cite{zong2025vl,chen2025true}.

Despite growing interest to multimodal in-context learning~\cite{awadalla2023openflamingo,NEURIPS2024_a0303731,laurenccon2024building}, Case II tasks remain insufficiently studied and inadequately addressed. 
While most existing MICL methods and evaluations are still centered on Case I settings~\cite{NEURIPS2024_deeb4d6b,doveh2024towards,jiang2024many}, some recent efforts have begun to consider Case II problems. However, as shown in Figure ~\ref{fig:insight}(a), these approaches often treat Case II as a straightforward extension of generic MICL rather than as a distinct regime with unique challenges~\cite{NEURIPS2024_12d3e63b,gupta2025meta,chen2025true}. 
They primarily emphasize general improvements such as representation alignment or prompting strategies, while overlooking the context‑driven rule induction and abstraction. 
In this paper, we argue that explicitly inducing the underlying rules from contextual demonstrations is essential for solving Case II scenarios. Accordingly, we focus on Case II multimodal in‑context learning and emphasize developing models capable of inferring shared underlying rules from demonstration examples and reliably applying these rules to new inputs, as illustrated in Figure~\ref{fig:insight}(b). 

With recent advances in multimodal large language models (MLLMs), which exhibit strong reasoning and abstraction capabilities across vision–language tasks~\cite{bai2025qwen25vltechnicalreport}, it is natural to leverage MLLMs to explicitly induce and articulate underlying rules, thereby enabling more effective generalization and improved performance on Case II tasks.

As a result, we propose ERIM (Explicit Rule‑Induction for Multimodal In‑Context Learning), a unified framework that systematically equips multimodal models with the ability to explicitly induce, reason over, and apply underlying rules in Case II MICL scenarios. ERIM transforms implicit contextual cues in MICL into an explicit and learnable inductive process through a tightly coupled data construction and training paradigm.
Specifically, ERIM organizes multimodal data into rule‑centric MICL episodes and introduces explicit rule‑induction supervision, enabling models to learn how underlying rules are inferred from contextual demonstrations and how these rules govern query prediction. Building on such rule‑centric datasets, ERIM fine‑tunes multimodal large language models using a two‑stage training strategy that combines mixed cold‑start supervised fine‑tuning with reinforcement learning, resulting in stronger rule reasoning, improved generalization, and more stable performance on Case II tasks.

We conduct extensive experiments across diverse Case II MICL benchmarks, demonstrating that models trained under ERIM achieve substantial and consistent performance improvements over strong baselines. Beyond accuracy gains, we perform in-depth analyses on rule consistency and generalization behavior, showing that explicit rule-induction supervision leads models to induce rules that are not only correct but also stable and transferable across contexts. These results provide empirical evidence that explicitly inducing the correct underlying rules is a key factor driving performance improvements in multimodal in-context learning. 

In summary, our contributions are as follows:

\begin{itemize}
    \item To the best of our knowledge, this work is the first to systematically explore explicit rule induction as a principled mechanism for enhancing Case II multimodal in-context learning, revealing that making underlying rules explicit enables models to induce correct rules and achieve stronger MICL performance.
    \item We propose ERIM, a unified framework that integrates data construction, rule induction supervision generation, and two-stage training to make contextual rule induction explicit, scalable, and learnable.
    \item We conduct comprehensive experiments and analyses that validate the effectiveness, generalization, and interpretability benefits of explicit rule induction in multimodal in-context learning.
\end{itemize}

\section{Related Work}

\subsection{Multimodal Large Language Models}

Modern MLLMs achieve strong few-shot and in-context learning abilities by combining large-scale interleaved image–text pretraining with architectural advances for multimodal fusion. Flamingo~\cite{NEURIPS2022_960a172b} establishes this paradigm by enabling flexible visual–textual conditioning and demonstrates state-of-the-art few-shot performance on a wide range of vision–language tasks. Subsequent works extend this framework through instruction tuning and data scaling, including Otter~\cite{11007678} and MMICL~\cite{ICLR2024_41128e5b}, which improve instruction following and multi-image prompt understanding, as well as systematic design studies that lead to efficient models such as Idefics and its successors~\cite{NEURIPS2024_a0303731,laurenccon2024building}.
Other efforts focus on expanding modality coverage and visual fidelity. LLaVA-OneVision~\cite{li2024llavaonevisioneasyvisualtask} unifies single-image, multi-image, and video understanding within a single model, while OtterHD~\cite{li2023otterhd} emphasizes high-resolution perception and fine-grained spatial reasoning. Meanwhile, strong compact language backbones, such as the Phi-3 and Phi-3.5 series~\cite{abdin2024phi-}, provide powerful reasoning capabilities. Despite these advances, most existing MLLMs are trained on broad interleaved data and evaluated on Case I settings.
As a result, performance often reflects strong zero-shot abilities rather than genuine cross-example abstraction. 

\subsection{Multimodal In-Context Learning}

Recent work on MICL mainly improves efficiency and stability by freezing pretrained MLLMs and modifying how demonstrations are encoded or utilized. AIM~\cite{Gao_Qiao_Wu_Wang_Cao_Li_2025} compresses multimodal demonstrations into compact latent representations to reduce visual token overhead, while LIVE~\cite{NEURIPS2024_12d3e63b} distills task-relevant information into learnable in-context vectors for efficient VQA. CATP~\cite{li2025catpcontextuallyadaptivetoken} further addresses redundancy in multimodal contexts by pruning uninformative visual tokens to improve both efficiency and robustness. To better leverage multimodal context, TrueMICL~\cite{chen2025true} highlights models’ over-reliance on textual patterns and proposes attention reallocation with a dedicated benchmark, while CAMA~\cite{li2025cama} modulates attention logits in a training-free manner to stabilize multimodal context utilization. 
Despite these advances, most existing methods largely overlook the inductive reasoning capacity of MLLMs themselves, limiting their effectiveness on Case II MICL scenarios that require deeper rule induction, abstraction, and generalization from context.

\begin{figure*}[h]
    \centering
    \includegraphics[width=\linewidth]{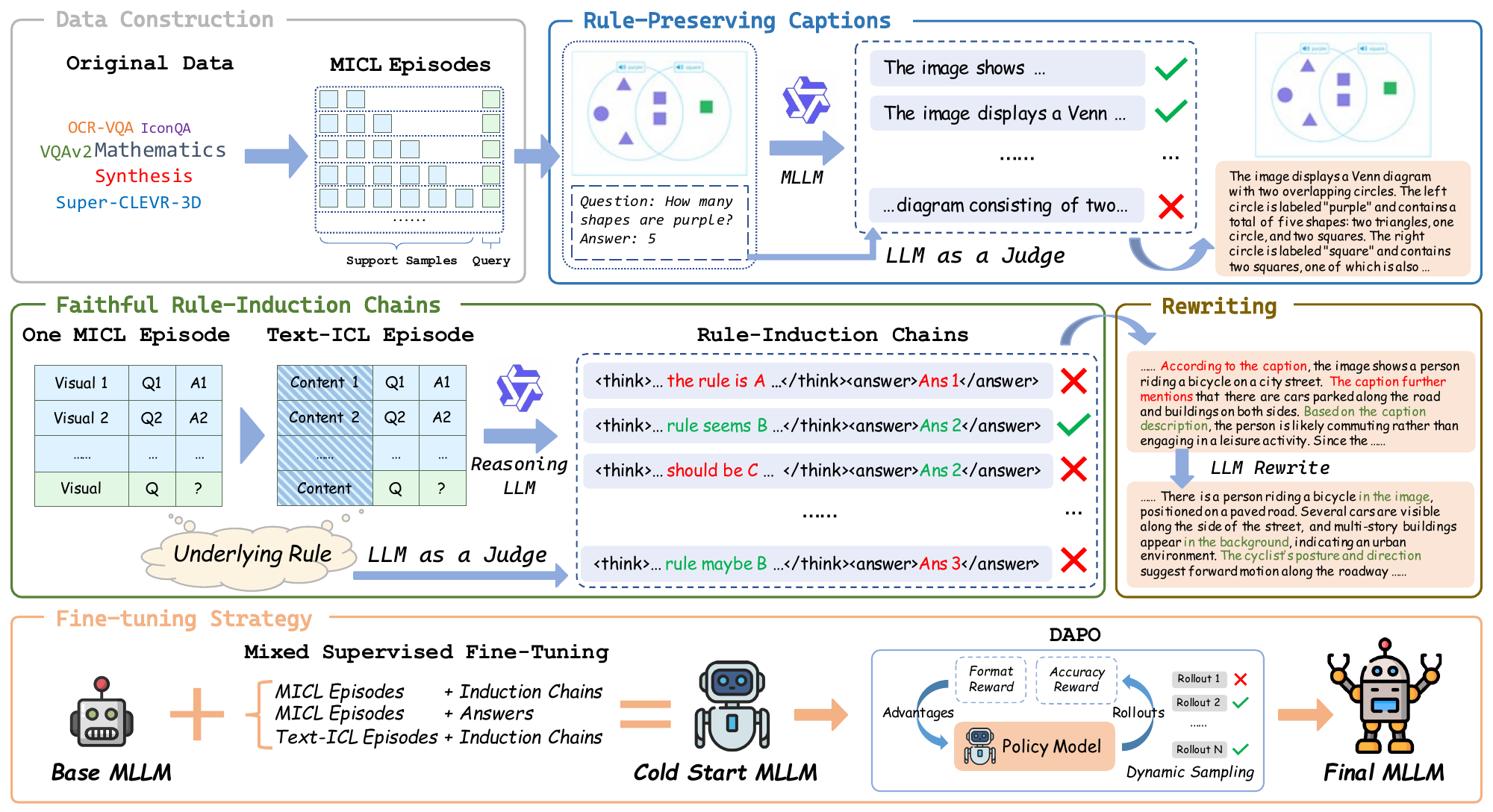}  
    \caption{Overview of ERIM for Case II multimodal in-context learning. ERIM organizes data into rule-centric MICL episodes and generates explicit rule-induction supervision by transforming multimodal episodes into text-based proxies with rule-preserving captions and faithfully filtered reasoning chains. The model is trained with a two-stage strategy that combines mixed supervised fine-tuning and reinforcement learning, enabling robust induction and application of underlying rules from context.}
    \label{fig:method}
\end{figure*}

\section{Method}

This section presents \textbf{ERIM}, Explicit Rule-Induction for Multimodal In-Context Learning, a unified framework for Case II MICL, as illustrated in Figure~\ref{fig:method}. 

\subsection{Problem Formulation and Scope}
\label{sec:formulation-scope}

We focus on Case II Multimodal In-Context Learning, where individual samples are ill-defined in isolation with an underlying rule shared by demonstration examples. 
Formally, each learning instance in MICL is organized as an \textbf{episode}, consisting of multiple VQA-style \textbf{samples}: a support set of $n$ samples,
\begin{equation}
    S=\{(V_i^s,Q_i^s,A_i^s)\}_{i=1}^n,
\end{equation}
where $V_i^s$, $Q_i^s$, and $A_i^s$ denote the visual input, question, and answer, respectively, along with a query sample $(V^q, Q^q)$ whose answer $A^q$ is withheld at inference time.
All samples within an episode are governed by a shared \textbf{underlying rule}, defining a mapping from visual inputs and question to the final answer.

Our goal is to fine-tune an MLLM that, given a support set and a query, can first \emph{explicitly induce the underlying rule} shared by the episode, and then apply the induced rule to predict the final answer. Concretely, the model is trained to perform rule induction conditioned on the support set,
\begin{equation}
R_{\mathrm{pred}} = \mathcal{M}(\mathrm{Prompt}; (V_1^s, Q_1^s, A_1^s), \dots, (V_n^s, Q_n^s, A_n^s)),
\end{equation}
where $R_{\mathrm{pred}}$ denotes the explicitly induced rule, and $\mathrm{Prompt}$ is the prompt template. The model then applies this rule to the query sample to produce the answer,
\begin{equation}
A_{\mathrm{pred}} = \mathcal{M}(\mathrm{Prompt}; R_{\mathrm{pred}}; (V^q, Q^q)).
\end{equation}
In practice, these two steps are implemented using a unified \emph{think-then-answer} prompt template, where the model is encouraged to first reason about and articulate the underlying rule before generating the final answer.

\subsection{Rule-Centric Data Construction}
\label{sec:rule-centric-data}

We construct a \textbf{Rule-Centric Multimodal In-Context Learning Dataset} by organizing samples around shared underlying rules, which define valid Case II MICL episodes.
The construction process starts from existing multimodal datasets and involves deriving rule definitions from annotations, grouping samples accordingly, and sampling support–query sets within each group. We apply this procedure to both typical VQA datasets and synthetic multimodal data, with rule construction adapted to the characteristics of each data source.




\subsubsection{Typical VQA Datasets}



For typical VQA datasets, underlying rules are not explicitly annotated and are derived from existing labels. We select datasets with sufficiently regular question structure for coherent rule construction,
including VQAv1~\cite{antol2015vqa}, VQAv2~\cite{goyal2017making}, IconQA~\cite{lu2021iconqa}, and OCR-VQA~\cite{mishra2019ocr}, excluding datasets that are unsuitable for coherent rule construction. We define underlying rules by grouping samples with identical question texts, 
where each unique question text specifies a shared rule that governs the visual–answer mapping. Additional processing details are provided in the Appendix~\ref{suppl:process-typical-VQA}.

\subsubsection{Synthetic Multimodal Data}



The detailed and accurate annotations of synthetic or rendered datasets enable more controlled underlying rule design.
For spatial perception and reasoning, we use Super-CLEVR-3D~\cite{wang20233d}, a challenging extension of CLEVR~\cite{johnson2017clevr}, and manually design underlying rules that require reasoning over 3D spatial configurations and object relations. For mathematical reasoning, we use rendered datasets of mathematical expressions and construct underlying rules based on simple arithmetic or symbolic relationships. 
More processing details are provided in the Appendix~\ref{suppl:rule-designs-for-synthetic-data}.

\subsection{Explicit Rule-Induction Supervision Generation}
\label{sec:rule-induction-supervision}

To enhance rule-induction in Case II MICL, we generate \textbf{Explicit Rule-Induction Supervision} from the Rule-Centric MICL Dataset. The core idea is to transform MICL episodes into text-only ICL tasks by replacing visual inputs with captions that preserve all rule-relevant information. This allows powerful language reasoning models to operate as proxies for MICL, producing explicit reasoning chains that reveal how the underlying rules are induced from the support set and applied to the query. 
\subsubsection{Generating Rule-Preserving Captions}

In Case II MICL, where underlying rules are constructed from specific visual attributes or relations, naively replacing images with captions may introduce severe and cumulative modality gaps, as any loss or distortion of rule-relevant visual information may invalidate the underlying rule of the episode and prevent correct rule induction.

To mitigate this issue, we adopt a caption filtering procedure to preserve critical visual details. For each image, we input the visual data together with the annotations used for rule construction into a MLLM to generate multiple candidate captions covering all relevant information. Each candidate is then paired with corresponding annotations and evaluated by another LLM, which judges whether the caption fully captures the annotated content. We retain only captions that fully capture the critical visual details, ensuring rule-preserving text representations for subsequent reasoning.

\subsubsection{Producing Faithful Rule-Induction Chains}



Using the rule-preserving captions, we replace all visual inputs in Rule-Centric MICL episodes, converting them into text-only in-context learning tasks. These episodes are then processed by a reasoning LLM, which performs rule induction and application based on the caption-based representations, acting as a proxy for MICL. We apply rejection sampling to retain only reasoning chains that are both correct and rule-consistent. 

Specifically, for each episode, the reasoning LLM generates multiple candidate reasoning chains. We first require the final answer enclosed in the \texttt{<answer> </answer>} tag to match the ground-truth query answer, ensuring answer-level correctness. We then evaluate whether the induced rule expressed in the reasoning chain is consistent with the underlying rule defined during episode construction by jointly providing the generated reasoning chain and the ground-truth rule description to another LLM for consistency judgment. 

\subsubsection{Mitigating Modality Drift by Rewriting}


Even after filtering for correct and rule-consistent rule-induction chains, some retained rule-induction chains may exhibit \emph{modality drift} in long-context reasoning, where the model gradually deviates from a multimodal reasoning role and instead explicitly refers to caption text. To address this issue, we use a LLM to refine the retained rule-induction chains by rewriting each chain to eliminate caption-specific references and correct inappropriate phrasing. Through this rewriting process, we obtain coherent, faithful, and multimodal-consistent rule-induction chains, which constitute the final \textbf{Explicit Rule-Induction Supervision}.




\subsection{Two-Stage Fine-tuning Strategy}
\label{sec:training-strategy}




\subsubsection{Mixed Cold-Start via Supervised Fine-Tuning}

We initialize the model using mixed cold-start supervised fine-tuning on a combination of three data forms derived from the same MICL episodes: original episodes with direct answer supervision, caption-based text-only episodes and original episodes with explicit rule-induction supervision. 
After this mixed supervised fine-tuning stage, we denote the resulting initialized model as $ \mathcal{M}_{\mathrm{SFT}} $. This model exhibits aligned output formats and basic rule-induction behaviors, providing a stable foundation for subsequent reinforcement learning.

\subsubsection{Reinforcement Learning with DAPO}

To further improve generalization beyond supervised reasoning chains, we apply reinforcement learning on rule-centric MICL episodes without explicit rule-induction supervision, denoted as $\mathcal{D}_{\mathrm{RL}}$. We adopt \textbf{DAPO}~\cite{yu2026dapo}, a variant of Group Relative Policy Optimization~\cite{shao2024deepseekmath} that applies a clip-higher operation to enable broader exploration and performs dynamic sampling to discard groups with uninformative (extremely high or low) rewards.


For each MICL prompt $x$, the current policy $ \pi_\theta $ which is initialized as $ \mathcal{M}_{\mathrm{SFT}} $ samples a group of $G$ candidate outputs $ \{\hat{y}_i\}_{i=1}^G $. Each output is assigned a reward:
\begin{equation}
    R(\hat{y}) = \lambda_{\mathrm{fmt}} R_{\mathrm{fmt}}(\hat{y}) + \lambda_{\mathrm{acc}} R_{\mathrm{acc}}(\hat{y}),
\end{equation}
where $ R_{\mathrm{fmt}} $ checks whether the output follows the required \texttt{<think>...</think><answer>...</answer>} format, and $ R_{\mathrm{acc}} $ indicates whether the extracted answer matches the ground truth. 
The resulting policy optimization objective is:
\begin{align}
J_{\mathrm{DAPO}}(\theta) & =\mathbb{E}_{x \sim \mathcal{D}_{\mathrm{RL}}}\mathbb{E}_{\left\{\hat{y}_i\right\}_{i=1}^G \sim \pi_{\theta_{\mathrm{old}}}(\cdot \mid x), \hat{y}_{i, t} \sim \hat{y}_{i}} \\
& \left[\operatorname{min} \left(\operatorname{clip}\left(r_{i,t}(\theta), 1-\varepsilon_{\mathrm{low}}, 1+\varepsilon_{\mathrm{high}}\right) A_i, \right.\right. \\
& \left.\left. r_{i,t}(\theta) A_{i}\right) - \beta D_{\mathrm{KL}}\left(\pi_{\theta} \| \pi_{\mathrm{ref}}\right)\right],
\end{align}
\begin{equation}
    r_{i,t}(\theta) = \frac{\pi_{\theta}\left(\hat{y}_i \mid x, \hat{y}_{i,<t}\right)}{\pi_{\theta_{\mathrm{old}}}\left(\hat{y}_i \mid x, \hat{y}_{i,<t}\right)}, A_i=\frac{R_i - \operatorname{mean}(\{R_i\})}{\operatorname{std}(\{R_i\})}
\end{equation}


\section{Experiments}

\subsection{Experimental Setup}

\definecolor{basecolor}{RGB}{153,153,153}
\definecolor{highlightcolor}{RGB}{225,225,225}

\begin{table*}[t]
\centering
{
\small
\setlength{\tabcolsep}{5pt}
\begin{tabular}{ll|ccccc|c||ccccc|c}
\toprule

\multicolumn{1}{c}{\multirow{3}{*}{Method}} &
\multicolumn{1}{c}{\multirow{3}{*}{Mode}} &
\multicolumn{6}{c}{CLEVR} & \multicolumn{6}{c}{Operator Induction} \\
\cmidrule(lr){3-8} \cmidrule(lr){9-14}
\multicolumn{2}{c}{} &
\multicolumn{1}{c}{2} & \multicolumn{1}{c}{3} & \multicolumn{1}{c}{4} &
\multicolumn{1}{c}{5} & \multicolumn{1}{c}{6} & \multicolumn{1}{c}{Avg} &
\multicolumn{1}{c}{2} & \multicolumn{1}{c}{3} & \multicolumn{1}{c}{4} &
\multicolumn{1}{c}{5} & \multicolumn{1}{c}{6} & \multicolumn{1}{c}{Avg} \\
\midrule
OpenFlamingo-9B & \textcolor{basecolor}{\textit{answer}} & 24.00 & 24.67 & 24.17 & 26.67 & 24.83 & 24.87 & 8.33 & 7.78 & 6.11 & 5.00 & 8.33 & 7.11 \\
IDEFICS3-8B-Llama3 & \textcolor{basecolor}{\textit{answer}} & 57.00 & 54.00 & 54.50 & 57.83 & 60.83 & 56.83 & 0.00 & 0.00 & 1.11 & 1.67 & 1.67 & 0.89 \\
Phi-3.5-Vision & \textcolor{basecolor}{\textit{answer}} & 32.17 & 26.83 & 28.33 & 26.00 & 28.33 & 28.33 & 10.56 & 15.56 & 11.11 & 6.11 & 7.78 & 10.22 \\
LLaVA-OneVision-7B & \textcolor{basecolor}{\textit{answer}} & 46.33 & 48.83 & 51.17 & 50.50 & 49.83 & 49.33 & 43.89 & 50.56 & 54.44 & 49.44 & 55.00 & 50.67 \\
OpenVLThinker-3B & \textit{reasoning} & 56.33 & 57.17 & 53.00 & 55.83 & 53.33 & 55.13 & 73.33 & 82.22 & 80.00 & 79.44 & 82.22 & 79.44 \\
VLAA-Thinker-3B & \textit{reasoning} & 56.83 & 56.67 & 54.67 & 51.50 & 48.67 & 53.67 & 67.78 & 76.11 & 83.33 & 83.89 & 80.56 & 78.33 \\
\rowcolor{highlightcolor} Qwen2.5-VL-3B & \textcolor{basecolor}{\textit{answer}} & 51.00 & 52.67 & 53.33 & 54.67 & 55.17 & 53.37 & 67.78 & 74.44 & 74.44 & 80.56 & 76.67 & 74.78 \\
ERIM-3B (Ours) & \textbf{\textit{induction}} & \textbf{64.00} & \textbf{64.33} & \textbf{65.33} & \textbf{66.83} & \textbf{63.83} & \textbf{64.87} & \textbf{88.89} & \textbf{92.22} & \textbf{90.56} & \textbf{96.67} & \textbf{97.22} & \textbf{93.11} \\
\midrule

\multicolumn{1}{c}{\multirow{3}{*}{Method}} & \multicolumn{1}{c}{\multirow{3}{*}{Mode}} &
\multicolumn{6}{c}{Matching MI} & \multicolumn{6}{c}{Clock} \\
\cmidrule(lr){3-8} \cmidrule(lr){9-14}
\multicolumn{2}{c}{} &
\multicolumn{1}{c}{0} & \multicolumn{1}{c}{1} & \multicolumn{1}{c}{2} &
\multicolumn{1}{c}{4} & \multicolumn{1}{c}{5} & \multicolumn{1}{c}{Avg} &
\multicolumn{1}{c}{2} & \multicolumn{1}{c}{3} & \multicolumn{1}{c}{4} &
\multicolumn{1}{c}{5} & \multicolumn{1}{c}{6} & \multicolumn{1}{c}{Avg} \\
\midrule
OpenFlamingo-9B & \textcolor{basecolor}{\textit{answer}} & 32.08 & 50.00 & 50.00 & 50.00 & 50.00 & 46.42 & 5.00 & 7.00 & 3.00 & 3.00 & 5.00 & 4.60 \\
IDEFICS3-8B-Llama3 & \textcolor{basecolor}{\textit{answer}} & 62.58 & 50.17 & 50.08 & 49.75 & 50.17 & 52.55 & 3.00 & 3.00 & 5.00 & 8.00 & 6.00 & 5.00 \\
Phi-3.5-Vision & \textcolor{basecolor}{\textit{answer}} & 68.42 & 49.75 & 49.92 & 50.08 & 50.08 & 53.65 & 7.00 & 5.00 & 11.00 & 6.00 & 6.00 & 7.00 \\
LLaVA-OneVision-7B & \textcolor{basecolor}{\textit{answer}} & \textbf{84.00} & 55.83 & 56.67 & 56.75 & 54.58 & 61.57 & 21.00 & 14.00 & 18.00 & 16.00 & 14.00 & 16.60 \\
OpenVLThinker-3B & \textit{reasoning} & 77.58 & 56.75 & 55.83 & 58.00 & 56.83 & 61.00 & 44.00 & 61.00 & 74.00 & 71.00 & \textbf{74.00} & 64.80 \\
VLAA-Thinker-3B & \textit{reasoning} & 75.50 & 60.67 & 62.50 & 63.08 & 65.33 & 65.42 & 42.00 & 52.00 & 52.00 & 60.00 & 54.00 & 52.00 \\
\rowcolor{highlightcolor} Qwen2.5-VL-3B & \textcolor{basecolor}{\textit{answer}} & 67.67 & 52.08 & 52.17 & 53.42 & 54.50 & 55.97 & 32.00 & 41.00 & 45.00 & 51.00 & 49.00 & 43.60 \\
ERIM-3B (Ours) & \textbf{\textit{induction}} & 64.17 & \textbf{76.25} & \textbf{77.08} & \textbf{81.58} & \textbf{82.08} & \textbf{76.23} & \textbf{78.00} & \textbf{71.00} & \textbf{83.00} & \textbf{77.00} & 71.00 & \textbf{76.00} \\
\midrule

\multicolumn{1}{c}{\multirow{3}{*}{Method}} & \multicolumn{1}{c}{\multirow{3}{*}{Mode}} &
\multicolumn{6}{c}{CLEVR (TrueMICL)} & \multicolumn{6}{c}{\textit{Subset-Level Average}} \\
\cmidrule(lr){3-8} \cmidrule(lr){9-14}
\multicolumn{2}{c}{} &
\multicolumn{1}{c}{2} & \multicolumn{1}{c}{3} & \multicolumn{1}{c}{4} &
\multicolumn{1}{c}{5} & \multicolumn{1}{c}{6} & \multicolumn{1}{c}{Avg} &
\multicolumn{1}{c}{/} & \multicolumn{1}{c}{/} & \multicolumn{1}{c}{/} &
\multicolumn{1}{c}{/} & \multicolumn{1}{c}{/} & \multicolumn{1}{c}{Avg} \\
\midrule
OpenFlamingo-9B & \textcolor{basecolor}{\textit{answer}} & 30.00 & 34.00 & 34.00 & 33.00 & 30.00 & 32.20 & 14.08 & 17.90 & 22.34 & 26.52 & 26.97 & 21.56 \\
IDEFICS3-8B-Llama3 & \textcolor{basecolor}{\textit{answer}} & \textbf{35.00} & 31.00 & 39.00 & 31.00 & \textbf{39.00} & 35.00 & 24.70 & 23.45 & 25.47 & 28.43 & 30.82 & 26.57 \\
Phi-3.5-Vision & \textcolor{basecolor}{\textit{answer}} & 25.00 & 29.00 & 29.00 & 35.00 & 30.00 & 29.60 & 20.38 & 17.73 & 21.63 & 20.83 & 21.45 & 20.40 \\
LLaVA-OneVision-7B & \textcolor{basecolor}{\textit{answer}} & 30.00 & 32.00 & 38.00 & 35.00 & 37.00 & 34.40 & 44.03 & 44.98 & 54.31 & 53.98 & 54.23 & 50.31 \\
OpenVLThinker-3B & \textit{reasoning} & 16.00 & 18.00 & 15.00 & 16.00 & 18.00 & 16.60 & 49.85 & 52.01 & 54.96 & 56.84 & 57.34 & 54.20 \\
VLAA-Thinker-3B & \textit{reasoning} & 11.00 & 12.00 & 6.00 & 12.00 & 12.00 & 10.60 & 44.01 & 45.95 & 48.29 & 52.61 & 53.03 & 48.78 \\
\rowcolor{highlightcolor} Qwen2.5-VL-3B & \textcolor{basecolor}{\textit{answer}} & 12.00 & 19.00 & 24.00 & 27.00 & 31.00 & 22.60 & 40.74 & 43.37 & 52.83 & 55.48 & 56.12 & 49.71 \\
ERIM-3B (Ours) & \textbf{\textit{induction}} & \textbf{35.00} & \textbf{37.00} & \textbf{46.00} & \textbf{44.00} & 32.00 & \textbf{38.80} & \textbf{58.69} & \textbf{71.91} & \textbf{77.36} & \textbf{78.02} & \textbf{75.40} & \textbf{72.28} \\
\bottomrule
\end{tabular}
}
\caption{Main results on Case II MICL benchmarks. We report accuracy (\%) of different methods across benchmark subsets under varying n-shot settings, along with the average performance over n-shots. For subsets with benchmark-specific additional information, results are averaged over all such settings. Due to space constraints, only 5 out of the 8 selected subsets are shown; the final block reports \textit{subset-level averages} to assess overall performance on Case II MICL benchmarks. 
}
\label{tab:main}
\end{table*}


\begin{figure}[h]
    \centering
    \includegraphics[width=0.96\linewidth]{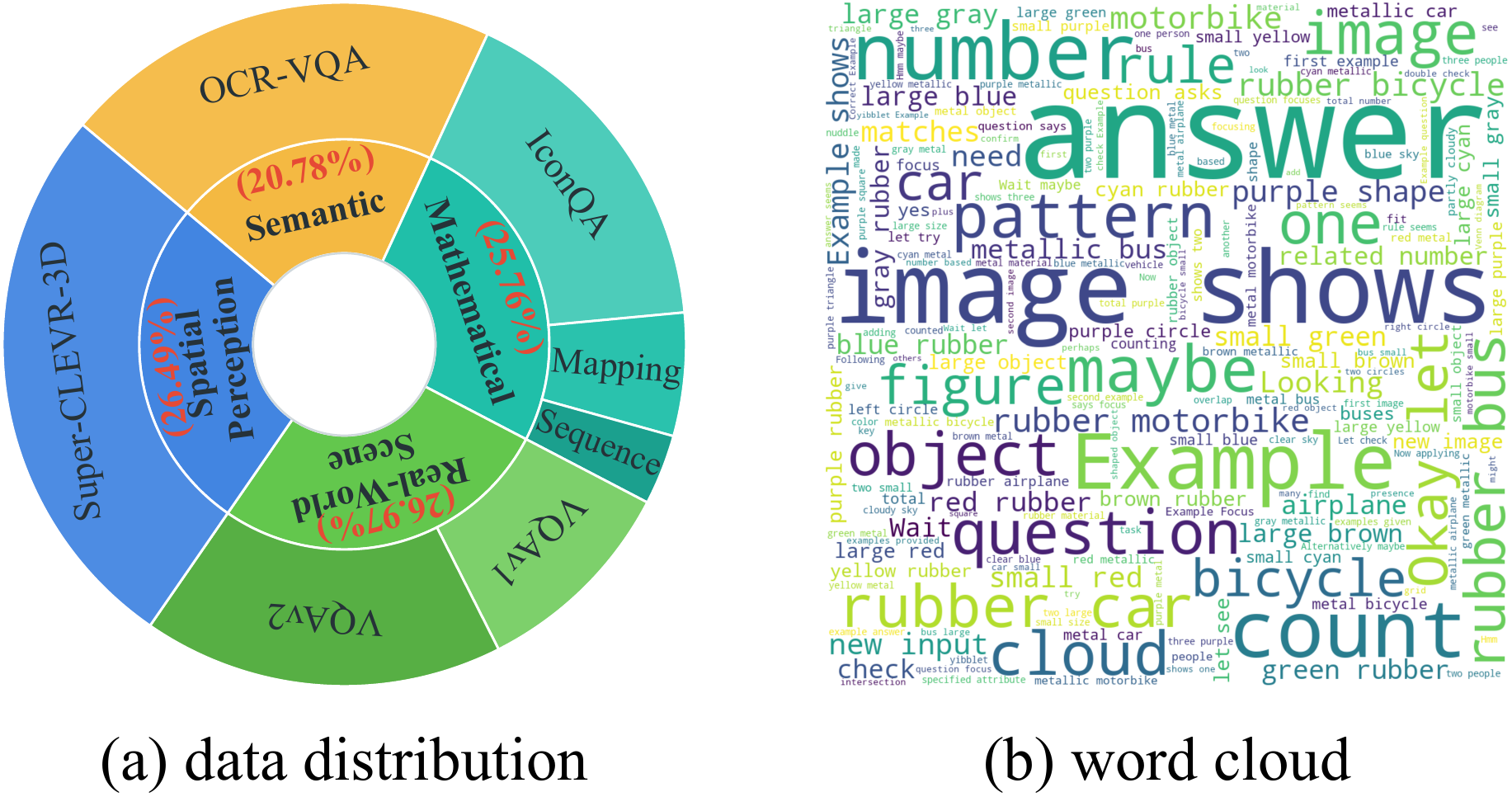}
    \caption{Statistical overview of the two core datasets used in our method.}
    \label{fig:statistic}
\end{figure}

\begin{table*}[t]
\centering
{
\small
\setlength{\tabcolsep}{5pt}
\begin{tabular}{lcc|ccccc|c||ccccc|c}
\toprule
\multicolumn{1}{c}{\multirow{2}{*}{Method}} & \multicolumn{2}{c}{Strategy} &
\multicolumn{6}{c}{CLEVR} & \multicolumn{6}{c}{Interleaved Operator Induction} \\
\cmidrule(lr){2-3} \cmidrule(lr){4-9} \cmidrule(lr){10-15}
\multicolumn{1}{c}{} &
\multicolumn{1}{c}{SFT} & \multicolumn{1}{c}{RL} & 
\multicolumn{1}{c}{2} & \multicolumn{1}{c}{3} & \multicolumn{1}{c}{4} &
\multicolumn{1}{c}{5} & \multicolumn{1}{c}{6} & \multicolumn{1}{c}{Avg} &
\multicolumn{1}{c}{2} & \multicolumn{1}{c}{3} & \multicolumn{1}{c}{4} &
\multicolumn{1}{c}{5} & \multicolumn{1}{c}{6} & \multicolumn{1}{c}{Avg} \\
\midrule
Qwen2.5VL & \ding{55} & \ding{55} & 51.00 & 52.67 & 53.33 & 54.67 & 55.17 & 53.37 & 54.44 & 60.56 & 62.22 & 58.89 & 66.11 & 60.44 \\
+Direct-SFT & Direct & \ding{55} & 60.17 & 61.17 & 61.17 & 62.83 & 59.83 & 61.03 & 67.22 & 73.33 & 75.56 & 76.11 & 78.89 & 74.22 \\
+Think-RL & \ding{55} & \ding{51} & 49.00 & 51.17 & 51.33 & 50.67 & 51.00 & 50.63 & 68.89 & 73.89 & 76.67 & \textbf{81.11} & \textbf{84.44} & 77.00 \\
+ERIM & Mixed & \ding{51} & \textbf{64.00} & \textbf{64.33} & \textbf{65.33} & \textbf{66.83} & \textbf{63.83} & \textbf{64.87} & \textbf{84.44} & \textbf{82.78} & \textbf{87.22} & 79.44 & 83.89 & \textbf{83.56} \\
\midrule
\multicolumn{1}{c}{\multirow{2}{*}{Method}} & \multicolumn{2}{c}{Strategy} &
\multicolumn{6}{c}{Clock} & \multicolumn{6}{c}{Palindrome} \\
\cmidrule(lr){2-3} \cmidrule(lr){4-9} \cmidrule(lr){10-15}
\multicolumn{1}{c}{} &
\multicolumn{1}{c}{SFT} & \multicolumn{1}{c}{RL} & 
\multicolumn{1}{c}{2} & \multicolumn{1}{c}{3} & \multicolumn{1}{c}{4} &
\multicolumn{1}{c}{5} & \multicolumn{1}{c}{6} & \multicolumn{1}{c}{Avg} &
\multicolumn{1}{c}{2} & \multicolumn{1}{c}{3} & \multicolumn{1}{c}{4} &
\multicolumn{1}{c}{5} & \multicolumn{1}{c}{6} & \multicolumn{1}{c}{Avg} \\
\midrule
Qwen2.5VL & \ding{55} & \ding{55} & 32.00 & 41.00 & 45.00 & 51.00 & 49.00 & 43.60 & 41.00 & 35.00 & 30.00 & 28.00 & 27.00 & 32.20 \\
+Direct-SFT & Direct & \ding{55} & 31.00 & 44.00 & 37.00 & 46.00 & 43.00 & 40.20 & 39.00 & 34.00 & 37.00 & 33.00 & 32.00 & 35.00 \\
+Think-RL & \ding{55} & \ding{51} & 64.00 & \textbf{72.00} & 81.00 & \textbf{87.00} & \textbf{83.00} & \textbf{77.40} & \textbf{63.00} & 66.00 & 68.00 & 57.00 & 62.00 & 63.20 \\
+ERIM & Mixed & \ding{51} & \textbf{78.00} & 71.00 & \textbf{83.00} & 77.00 & 71.00 & 76.00 & 55.00 & \textbf{68.00} & \textbf{73.00} & \textbf{81.00} & \textbf{77.00} & \textbf{70.80} \\
\midrule
\multicolumn{1}{c}{\multirow{2}{*}{Method}} & \multicolumn{2}{c}{Strategy} &
\multicolumn{6}{c}{CLEVR (TrueMICL)} & \multicolumn{6}{c}{\textit{Subset-Level Average}} \\
\cmidrule(lr){2-3} \cmidrule(lr){4-9} \cmidrule(lr){10-15}
\multicolumn{1}{c}{} &
\multicolumn{1}{c}{SFT} & \multicolumn{1}{c}{RL} & 
\multicolumn{1}{c}{2} & \multicolumn{1}{c}{3} & \multicolumn{1}{c}{4} &
\multicolumn{1}{c}{5} & \multicolumn{1}{c}{6} & \multicolumn{1}{c}{Avg} &
\multicolumn{1}{c}{/} & \multicolumn{1}{c}{/} & \multicolumn{1}{c}{/} &
\multicolumn{1}{c}{/} & \multicolumn{1}{c}{/} & \multicolumn{1}{c}{Avg} \\
\midrule
Qwen2.5VL & \ding{55} & \ding{55} & 12.00 & 19.00 & 24.00 & 27.00 & 31.00 & 22.60 & 40.74 & 43.37 & 52.83 & 55.48 & 56.12 & 49.71 \\
+Direct-SFT & Direct & \ding{55} & 31.00 & 31.00 & 31.00 & 30.00 & 31.00 & 30.80 & 46.42 & 62.78 & 63.16 & 64.36 & 63.67 & 60.08 \\
+Think-RL & \ding{55} & \ding{51} & 23.00 & 24.00 & 22.00 & 19.00 & 21.00 & 21.80 & 54.25 & 69.41 & 72.26 & 72.18 & 72.73 & 68.17 \\
+ERIM & Mixed & \ding{51} & \textbf{35.00} & \textbf{37.00} & \textbf{46.00} & \textbf{44.00} & \textbf{32.00} & \textbf{38.80} & \textbf{58.69} & \textbf{71.91} & \textbf{77.36} & \textbf{78.02} & \textbf{75.40} & \textbf{72.28} \\
\bottomrule
\end{tabular}
}
\caption{Ablation results of training strategies on Qwen2.5-VL-3B-Instruct. We report accuracy (\%) with the same metrics and settings as Table~\ref{tab:main}. "+Direct-SFT" denotes supervised fine-tuning on the Rule-Centric MICL Dataset using the direct-answer prompt; "+Think-RL" denotes DAPO training with the think-then-answer prompt on the same dataset; "+ERIM" corresponds to the full proposed training strategy with mixed SFT and DAPO. Due to space constraints, only 5 of the 8 selected subsets are shown. Full results are provided in the Appendix~\ref{suppl:complete-ablation-strategy-results}.}
\label{tab:ablation-strategy}
\end{table*}

\begin{table}[ht]
\centering
{
\small
\setlength{\tabcolsep}{5pt}
\begin{tabular}{c|cccc}
\toprule
benchmark & zero-shot & few-shot & ERI prompt & ERIM-3B \\
\midrule
VL-ICL & 20.34 & 31.65 & 40.30 & 78.52 \\
TrueMICL & 26.60 & 30.93 & 40.33 & 61.87 \\
\bottomrule
\end{tabular}
}
\caption{Effect of prompting strategies. Zero-shot uses the basic thinking template, few-shot additionally provides rule-induction chains as demonstrations, and ERI prompt explicitly instructs the model to induce and apply the underlying rule. We report average accuracy (\%) on VL-ICL and TrueMICL, averaged over all other configurations.}
\label{tab:ablation-prompt}
\end{table}

\begin{figure*}[h]
    \centering
    \includegraphics[width=\textwidth]{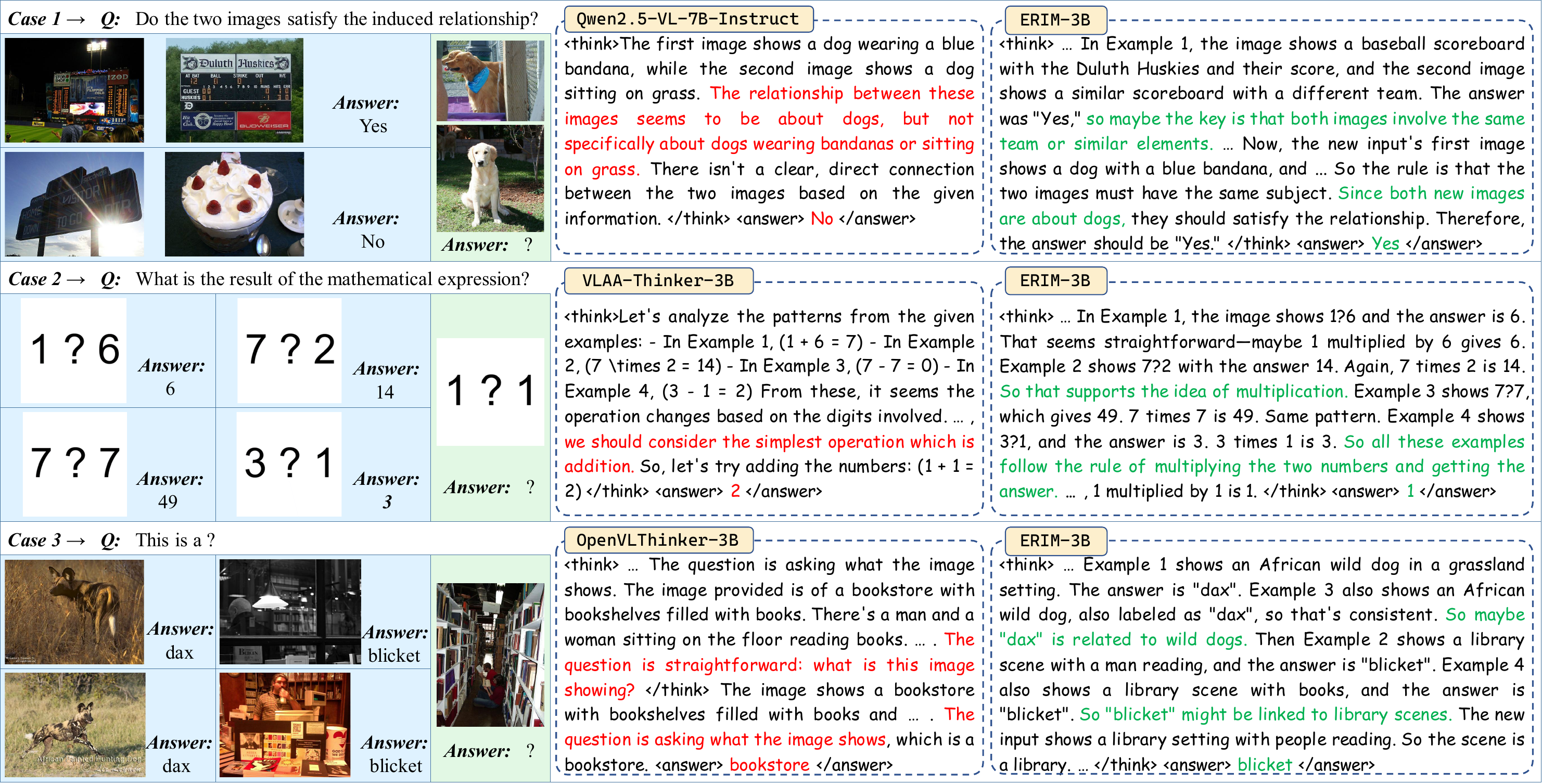}
    \caption{Qualitative case study. The blue region indicates the support set, and the green region indicates the query sample. We report the outputs of Qwen2.5-VL-7B-Instruct, VLAA-Thinker-3B, OpenVLThinker-3B, and our method for comparison. Incorrect reasoning steps are highlighted in red, while correct ones are highlighted in green.}
    \label{fig:case}
\end{figure*}

\begin{figure}[h]
    \centering
    \includegraphics[width=0.94\linewidth]{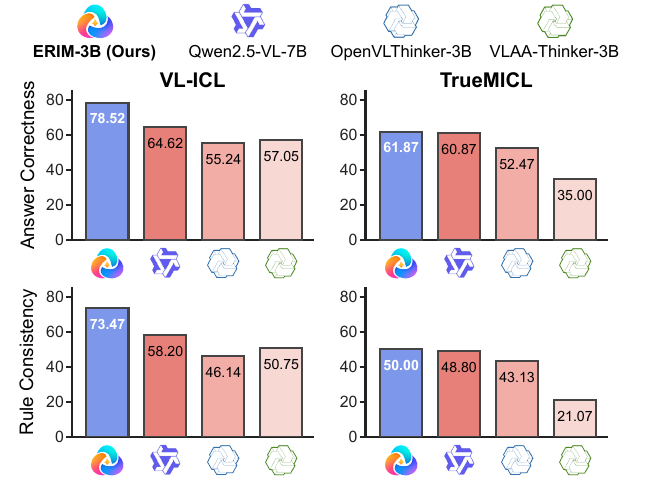}
    \caption{Comparison of answer correctness and rule consistency across four models. Rule consistency is evaluated using Qwen2.5-72B-Instruct as the LLM judge. We report average accuracy (\%) on VL-ICL and TrueMICL, averaged over all other configurations.}
    \label{fig:rule}
\end{figure}

\subsubsection{Benchmarks and Baselines}

We first select datasets from existing MICL benchmarks that satisfy the Case II setting, and then further filter out subsets that can already be well solved by Qwen2.5-VL-3B-Instruct~\cite{bai2025qwen25vltechnicalreport}. Following this process, we retain five subsets, \emph{open mi}, \emph{clevr}, \emph{operator induction}, \emph{interleaved operator induction}, and \emph{matching mi}, from VL-ICL~\cite{zong2025vl}, and three subsets, \emph{clock}, \emph{palindrome}, and \emph{clevr (TrueMICL)}, from TrueMICL~\cite{chen2025true}. These benchmarks span diverse rule types, including visual relations, symbolic operations, and abstract pattern induction.

We compare our method with representative MICL models of comparable scale that do not rely on task-specific modules or additional supervision. 
We include OpenFlamingo-9B~\cite{awadalla2023openflamingo} and IDEFICS3-8B-Llama3~\cite{laurenccon2024building} as representative MICL baselines. To further broaden the comparison, we consider Phi-3.5-Vision~\cite{abdin2024phi-}, LLaVA-OneVision-7B~\cite{li2024llava} and Qwen2.5-VL-3B-Instruct~\cite{bai2025qwen25vltechnicalreport}, which are general-purpose MLLMs with strong multimodal understanding capabilities. We further include OpenVLThinker-3B~\cite{deng2025openvlthinker} and VLAA-Thinker-3B~\cite{chen2025sft}, which are explicitly designed for multimodal reasoning, to provide a more comprehensive evaluation across different modeling paradigms.

\subsubsection{Dataset Overview}

Figure~\ref{fig:statistic} summarizes the two datasets used in our method. 
The left panel in Figure~\ref{fig:statistic}(a) presents the distribution of the \textbf{Rule-Centric MICL Dataset}, which is grouped into four categories by contextual induction ability.
The right panel in Figure~\ref{fig:statistic}(b) shows the word cloud of the generated \textbf{Explicit Rule-Induction Supervision}, indicating diverse and explicit reasoning patterns that provide an effective language and reasoning format for rule induction.

\subsubsection{Implementation}

In our implementation, the MLLM used for generating captions is \textbf{Qwen2.5-VL-72B-Instruct}~\cite{bai2025qwen25vltechnicalreport}. \textbf{Qwen2.5-72B-Instruct}~\cite{qwen2025qwen25technicalreport} serves as the LLM for judgment and rewriting. In addition, the reasoning LLM used to generate explicit rule-induction supervision is \textbf{QwQ-32B}~\cite{qwq32b}. We adopt \textbf{Qwen2.5-VL-3B-Instruct}~\cite{bai2025qwen25vltechnicalreport} as the base model to conduct experiments. Further details are provided in the Appendix~\ref{suppl:implementation-details}.
For the Case II MICL setting, we design a unified prompt template that flexibly accommodates benchmark-specific information, such as task descriptions required by different benchmarks.
The template supports a direct-answer mode and a think-then-answer mode that explicitly induces the underlying rule before prediction. 
All prompt templates involved in our method are provided in the Appendix~\ref{suppl:all-prompt-templates}.



\subsection{Main Results}

We denote the final model obtained after the two-stage training strategy as \textbf{ERIM-3B}. Table~\ref{tab:main} presents the main experimental results. ERIM-3B achieves the best performance on most reported subsets and obtains a clear lead in the subset-level average, demonstrating strong generalization across different visual domains and rule types. The gains are especially pronounced on CLEVR, Operator Induction, and Clock, indicating that explicit rule induction is critical for Case II scenarios.
Another notable trend is that ERIM-3B maintains strong performance under different n-shot settings, showing that the model can effectively leverage contextual examples in both relatively low-shot and higher-shot regimes. These results confirm the effectiveness of our method for explicit rule induction and application in Case II MICL.

\subsection{Ablation Studies}

\subsubsection{Effect of Training Strategy}

Table~\ref{tab:ablation-strategy} compares different training strategies on a fixed base model. Most variants outperform the vanilla base model, indicating the effectiveness of rule-centric MICL data. Direct-SFT brings limited and saturating gains, while Think-RL significantly improves performance on reasoning-intensive subsets but shows less stable improvements when used alone. Our full strategy consistently achieves the best results across most reported subsets and the highest subset-level average. This demonstrates that the two-stage training is essential: mixed SFT initialization stabilizes learning, and RL further strengthens explicit rule induction and generalization. 

\subsubsection{Effect of Prompting Strategy}

To examine whether the improvements of ERIM can be attributed solely to prompting, we compare several prompting strategies in Table~\ref{tab:ablation-prompt}. Explicitly requesting rule induction improves the zero-shot results, but remains far below ERIM-3B on both benchmarks. Providing high-quality rule-induction chains as few-shot demonstrations also yields only limited gains. These results indicate that neither explicit prompting nor in-context exposure to correct reasoning chains is sufficient to reliably elicit Case II MICL capabilities. Effective rule induction must instead be internalized through dedicated supervision and optimization.


\subsection{Analysis}

\subsubsection{Qualitative Case Study}

Figure~\ref{fig:case} presents three representative cases 
to illustrate how correct rule induction directly determines prediction accuracy.
In all cases, existing reasoning models fail to infer the underlying rule from the demonstrations and consequently produce incorrect answers. In contrast, ERIM-3B consistently identifies the shared relationship, recovers the hidden mathematical operation, or induces the correct mapping between visual concepts and nonsensical labels, leading to accurate predictions. These examples demonstrate that reliable rule induction is critical for solving Case II MICL tasks and further support our central insight that improving explicit rule-induction ability directly enhances final prediction accuracy.

\subsubsection{Rule Consistency Evaluation}

To further analyze whether correct rule induction leads to correct final predictions, we conduct a rule consistency evaluation across models. As shown in Figure \ref{fig:rule}, answer correctness and rule consistency exhibit highly aligned trends across both VL-ICL and TrueMICL. Models with more accurate induced rules consistently achieve higher answer accuracy, while weaker rule induction leads to degraded performance. 
This correlation provides quantitative evidence for our core insight: explicitly inducing more accurate rules is crucial for improving performance on Case II MICL tasks.

\section{Conclusion}

This paper studies MICL scenarios that require inducing underlying rules from multimodal demonstrations and emphasizes the role of explicit rule induction for robust generalization. We propose ERIM, a unified framework that constructs rule-centric MICL data, generates explicit rule-induction supervision, and trains MLLMs through a two-stage strategy. Extensive experiments and analyses demonstrate that ERIM consistently outperforms strong baselines, validating the effectiveness of explicit rule induction. 

\bibliography{aaai2027}

@inproceedings{zong2025vl,
  title={Vl-icl bench: The devil in the details of multimodal in-context learning},
  author={Zong, Yongshuo and Bohdal, Ondrej and Hospedales, Timothy},
  booktitle={International Conference on Learning Representations},
  volume={2025},
  pages={100058--100100},
  year={2025}
}

@article{chen2025true,
  title={True multimodal in-context learning needs attention to the visual context},
  author={Chen, Shuo and Liu, Jianzhe and Han, Zhen and Xia, Yan and Cremers, Daniel and Torr, Philip and Tresp, Volker and Gu, Jindong},
  journal={arXiv preprint arXiv:2507.15807},
  year={2025}
}

@inproceedings{antol2015vqa,
  title={Vqa: Visual question answering},
  author={Antol, Stanislaw and Agrawal, Aishwarya and Lu, Jiasen and Mitchell, Margaret and Batra, Dhruv and Zitnick, C Lawrence and Parikh, Devi},
  booktitle={Proceedings of the IEEE international conference on computer vision},
  pages={2425--2433},
  year={2015}
}

@inproceedings{goyal2017making,
  title={Making the v in vqa matter: Elevating the role of image understanding in visual question answering},
  author={Goyal, Yash and Khot, Tejas and Summers-Stay, Douglas and Batra, Dhruv and Parikh, Devi},
  booktitle={Proceedings of the IEEE conference on computer vision and pattern recognition},
  pages={6904--6913},
  year={2017}
}

@article{lu2021iconqa,
  title={Iconqa: A new benchmark for abstract diagram understanding and visual language reasoning},
  author={Lu, Pan and Qiu, Liang and Chen, Jiaqi and Xia, Tony and Zhao, Yizhou and Zhang, Wei and Yu, Zhou and Liang, Xiaodan and Zhu, Song-Chun},
  journal={arXiv preprint arXiv:2110.13214},
  year={2021}
}

@inproceedings{mishra2019ocr,
  title={Ocr-vqa: Visual question answering by reading text in images},
  author={Mishra, Anand and Shekhar, Shashank and Singh, Ajeet Kumar and Chakraborty, Anirban},
  booktitle={2019 international conference on document analysis and recognition (ICDAR)},
  pages={947--952},
  year={2019},
  organization={IEEE}
}

@inproceedings{johnson2017clevr,
  title={Clevr: A diagnostic dataset for compositional language and elementary visual reasoning},
  author={Johnson, Justin and Hariharan, Bharath and Van Der Maaten, Laurens and Fei-Fei, Li and Lawrence Zitnick, C and Girshick, Ross},
  booktitle={Proceedings of the IEEE conference on computer vision and pattern recognition},
  pages={2901--2910},
  year={2017}
}

@article{wang20233d,
  title={3d-aware visual question answering about parts, poses and occlusions},
  author={Wang, Xingrui and Ma, Wufei and Li, Zhuowan and Kortylewski, Adam and Yuille, Alan L},
  journal={Advances in Neural Information Processing Systems},
  volume={36},
  pages={58717--58735},
  year={2023}
}

@misc{bai2025qwen25vltechnicalreport,
      title={Qwen2.5-VL Technical Report}, 
      author={Shuai Bai and Keqin Chen and Xuejing Liu and Jialin Wang and Wenbin Ge and Sibo Song and Kai Dang and Peng Wang and Shijie Wang and Jun Tang and Humen Zhong and Yuanzhi Zhu and Mingkun Yang and Zhaohai Li and Jianqiang Wan and Pengfei Wang and Wei Ding and Zheren Fu and Yiheng Xu and Jiabo Ye and Xi Zhang and Tianbao Xie and Zesen Cheng and Hang Zhang and Zhibo Yang and Haiyang Xu and Junyang Lin},
      year={2025},
      eprint={2502.13923},
      archivePrefix={arXiv},
      primaryClass={cs.CV},
      url={https://arxiv.org/abs/2502.13923}, 
}

@article{awadalla2023openflamingo,
  title={Openflamingo: An open-source framework for training large autoregressive vision-language models},
  author={Awadalla, Anas and Gao, Irena and Gardner, Josh and Hessel, Jack and Hanafy, Yusuf and Zhu, Wanrong and Marathe, Kalyani and Bitton, Yonatan and Gadre, Samir and Sagawa, Shiori and others},
  journal={arXiv preprint arXiv:2308.01390},
  year={2023}
}

@article{laurenccon2024building,
  title={Building and better understanding vision-language models: insights and future directions},
  author={Lauren{\c{c}}on, Hugo and Marafioti, Andr{\'e}s and Sanh, Victor and Tronchon, L{\'e}o},
  journal={arXiv preprint arXiv:2408.12637},
  year={2024}
}

@techreport{abdin2024phi-,
author = {Abdin, Marah I and Ade Jacobs, Sam and Awan, Ammar Ahmad and Aneja, Jyoti and Awadallah, Ahmed and Hassan Awadalla, Hany and Bach, Nguyen and Bahree, Amit and Bakhtiari, Arash and Behl, Harkirat and Benhaim, Alon and Bilenko, Misha and Bjorck, Johan and Bubeck, Sébastien and Cai, Martin and Mendes, Caio César Teodoro and Chen, Weizhu and Chaudhary, Vishrav and Chopra, Parul and Giorno, Allie Del and de Rosa, Gustavo and Dixon, Matthew and Eldan, Ronen and Iter, Dan and Goswami, Abhishek and Gunasekar, Suriya and Haider, Emman and Hao, Junheng and Russell J. Hewett and Huynh, Jamie and Javaheripi, Mojan and Jin, Xin and Kauffmann, Piero and Karampatziakis, Nikos and Kim, Dongwoo and Khademi, Mahmoud and Kurilenko, Lev and Lee, James R. and Lee, Yin Tat and Li, Yuanzhi and Liang, Chen and Liu, Weishung and Lin, Xihui (Eric) and Lin, Zeqi and Madan, Piyush and Mitra, Arindam and Modi, Hardik and Nguyen, Anh and Norick, Brandon and Patra, Barun and Perez-Becker, Daniel and Portet, Thomas and Pryzant, Reid and Qin, Heyang and Radmilac, Marko and Rosset, Corby and Roy, Sambudha and Saarikivi, Olli and Saied, Amin and Salim, Adil and Santacroce, Michael and Shah, Shital and Shang, Ning and Sharma, Hiteshi and Song, Xia and Ruwase, Olatunji and Wang, Xin and Ward, Rachel and Wang, Guanhua and Witte, Philipp and Wyatt, Michael and Xu, Can and Xu, Jiahang and Xu, Weijian and Yadav, Sonali and Yang, Fan and Yang, Ziyi and Yu, Donghan and Zhang, Chengruidong and Zhang, Cyril and Zhang, Jianwen and Zhang, Li Lyna and Zhang, Yi and Zhang, Yunan and Zhou, Xiren},
title = {Phi-3 Technical Report: A Highly Capable Language Model Locally on Your Phone},
institution = {Microsoft},
year = {2024},
month = {August},
url = {https://www.microsoft.com/en-us/research/publication/phi-3-technical-report-a-highly-capable-language-model-locally-on-your-phone/},
number = {MSR-TR-2024-12},
}

@article{li2024llava,
  title={Llava-onevision: Easy visual task transfer},
  author={Li, Bo and Zhang, Yuanhan and Guo, Dong and Zhang, Renrui and Li, Feng and Zhang, Hao and Zhang, Kaichen and Zhang, Peiyuan and Li, Yanwei and Liu, Ziwei and others},
  journal={arXiv preprint arXiv:2408.03326},
  year={2024}
}

@article{deng2025openvlthinker,
  title={Openvlthinker: An early exploration to complex vision-language reasoning via iterative self-improvement},
  author={Deng, Yihe and Bansal, Hritik and Yin, Fan and Peng, Nanyun and Wang, Wei and Chang, Kai-Wei},
  journal={arXiv preprint arXiv:2503.17352},
  year={2025}
}

@article{chen2025sft,
  title={Sft or rl? an early investigation into training r1-like reasoning large vision-language models},
  author={Chen, Hardy and Tu, Haoqin and Wang, Fali and Liu, Hui and Tang, Xianfeng and Du, Xinya and Zhou, Yuyin and Xie, Cihang},
  journal={arXiv preprint arXiv:2504.11468},
  year={2025}
}

@misc{qwq32b,
    title = {QwQ-32B: Embracing the Power of Reinforcement Learning},
    url = {https://qwenlm.github.io/blog/qwq-32b/},
    author = {Qwen Team},
    month = {March},
    year = {2025}
}

@misc{qwen2025qwen25technicalreport,
      title={Qwen2.5 Technical Report}, 
      author={Qwen and : and An Yang and Baosong Yang and Beichen Zhang and Binyuan Hui and Bo Zheng and Bowen Yu and Chengyuan Li and Dayiheng Liu and Fei Huang and Haoran Wei and Huan Lin and Jian Yang and Jianhong Tu and Jianwei Zhang and Jianxin Yang and Jiaxi Yang and Jingren Zhou and Junyang Lin and Kai Dang and Keming Lu and Keqin Bao and Kexin Yang and Le Yu and Mei Li and Mingfeng Xue and Pei Zhang and Qin Zhu and Rui Men and Runji Lin and Tianhao Li and Tianyi Tang and Tingyu Xia and Xingzhang Ren and Xuancheng Ren and Yang Fan and Yang Su and Yichang Zhang and Yu Wan and Yuqiong Liu and Zeyu Cui and Zhenru Zhang and Zihan Qiu},
      year={2025},
      eprint={2412.15115},
      archivePrefix={arXiv},
      primaryClass={cs.CL},
      url={https://arxiv.org/abs/2412.15115}, 
}

@article{shao2024deepseekmath,
  title={Deepseekmath: Pushing the limits of mathematical reasoning in open language models},
  author={Shao, Zhihong and Wang, Peiyi and Zhu, Qihao and Xu, Runxin and Song, Junxiao and Bi, Xiao and Zhang, Haowei and Zhang, Mingchuan and Li, YK and Wu, Yang and others},
  journal={arXiv preprint arXiv:2402.03300},
  year={2024}
}

@article{yu2026dapo,
  title={Dapo: An open-source llm reinforcement learning system at scale},
  author={Yu, Qiying and Zhang, Zheng and Zhu, Ruofei and Yuan, Yufeng and Zuo, Xiaochen and Yue, Yu and Dai, Weinan and Fan, Tiantian and Liu, Gaohong and Liu, Lingjun and others},
  journal={Advances in Neural Information Processing Systems},
  volume={38},
  pages={113222--113244},
  year={2026}
}

@inproceedings{NEURIPS2022_960a172b,
 author = {Alayrac, Jean-Baptiste and Donahue, Jeff and Luc, Pauline and Miech, Antoine and Barr, Iain and Hasson, Yana and Lenc, Karel and Mensch, Arthur and Millican, Katherine and Reynolds, Malcolm and Ring, Roman and Rutherford, Eliza and Cabi, Serkan and Han, Tengda and Gong, Zhitao and Samangooei, Sina and Monteiro, Marianne and Menick, Jacob L and Borgeaud, Sebastian and Brock, Andy and Nematzadeh, Aida and Sharifzadeh, Sahand and Bi\'{n}kowski, Miko\l aj and Barreira, Ricardo and Vinyals, Oriol and Zisserman, Andrew and Simonyan, Kar\'{e}n},
 booktitle = {Advances in Neural Information Processing Systems},
 editor = {S. Koyejo and S. Mohamed and A. Agarwal and D. Belgrave and K. Cho and A. Oh},
 pages = {23716--23736},
 publisher = {Curran Associates, Inc.},
 title = {Flamingo: a Visual Language Model for Few-Shot Learning},
 url = {https://proceedings.neurips.cc/paper_files/paper/2022/file/960a172bc7fbf0177ccccbb411a7d800-Paper-Conference.pdf},
 volume = {35},
 year = {2022}
}

@ARTICLE{11007678,
  author={Li, Bo and Zhang, Yuanhan and Chen, Liangyu and Wang, Jinghao and Pu, Fanyi and Cahyono, Joshua Adrian and Yang, Jingkang and Li, Chunyuan and Liu, Ziwei},
  journal={IEEE Transactions on Pattern Analysis and Machine Intelligence}, 
  title={Otter: A Multi-Modal Model With In-Context Instruction Tuning}, 
  year={2025},
  volume={47},
  number={9},
  pages={7543-7557},
  doi={10.1109/TPAMI.2025.3571946}}

@inproceedings{ICLR2024_41128e5b,
 author = {Zhao, Haozhe and Cai, Zefan and Si, Shuzheng and Ma, Xiaojian and An, Kaikai and Chen, Liang and Liu, Zixuan and Wang, Sheng and Han, Wenjuan and Chang, Baobao},
 booktitle = {International Conference on Representation Learning},
 editor = {B. Kim and Y. Yue and S. Chaudhuri and K. Fragkiadaki and M. Khan and Y. Sun},
 pages = {14942--14980},
 title = {MMICL: Empowering Vision-language Model with Multi-Modal In-Context Learning},
 url = {https://proceedings.iclr.cc/paper_files/paper/2024/file/41128e5b3a7622da5b17588757599077-Paper-Conference.pdf},
 volume = {2024},
 year = {2024}
}

@misc{li2024llavaonevisioneasyvisualtask,
      title={LLaVA-OneVision: Easy Visual Task Transfer}, 
      author={Bo Li and Yuanhan Zhang and Dong Guo and Renrui Zhang and Feng Li and Hao Zhang and Kaichen Zhang and Peiyuan Zhang and Yanwei Li and Ziwei Liu and Chunyuan Li},
      year={2024},
      eprint={2408.03326},
      archivePrefix={arXiv},
      primaryClass={cs.CV},
      url={https://arxiv.org/abs/2408.03326}, 
}

@article{li2023otterhd,
  title={Otterhd: A high-resolution multi-modality model},
  author={Li, Bo and Zhang, Peiyuan and Yang, Jingkang and Zhang, Yuanhan and Pu, Fanyi and Liu, Ziwei},
  journal={arXiv preprint arXiv:2311.04219},
  year={2023}
}

@article{Gao_Qiao_Wu_Wang_Cao_Li_2025, 
  title={AIM: Let Any Multimodal Large Language Models Embrace Efficient In-Context Learning}, volume={39},
  url={https://ojs.aaai.org/index.php/AAAI/article/view/32316},
  DOI={10.1609/aaai.v39i3.32316},
  abstractNote={In-context learning (ICL) advances Large Language Models (LLMs) exhibiting emergent ability on downstream tasks without updating billions of parameters. However, in the area of multimodal Large Language Models (MLLMs), two problems hinder the application of multimodal ICL: (1) Most primary MLLMs are only trained on single-image datasets, making them unable to read extra multimodal demonstrations. (2) With the demonstrations increasing, thousands of visual tokens highly challenge hardware and degrade ICL performance. During preliminary explorations, we discovered that the inner LLM focuses more on the linguistic modality within multimodal demonstrations during generation. Therefore, we propose a general and lightweight framework AIM to tackle the mentioned problems through Aggregating Image information of Multimodal demonstrations to the latent space of the corresponding textual labels. After aggregation, AIM substitutes each demonstration with generated fused virtual tokens whose length is reduced to the same as its texts. Except for shortening input length, AIM further upgrades MLLMs pre-trained on image-text pairs to support multimodal ICL, as images from demonstrations are disregarded. Furthermore, benefiting from aggregating different demonstrations independently, AIM configures Demonstration Bank (DB) to avoid repeated aggregation, which significantly boosts model efficiency. We build AIM upon QWen-VL and LLaVA-Next, and AIM is comprehensively evaluated on image caption, VQA, and hateful speech detection. Outstanding results reveal that AIM provides an efficient and effective solution in upgrading MLLMs for multimodal ICL.},
  number={3},
  journal={Proceedings of the AAAI Conference on Artificial Intelligence},
  author={Gao, Jun and Qiao, Qian and Wu, Tianxiang and Wang, Zili and Cao, Ziqiang and Li, Wenjie},
  year={2025},
  month={Apr.},
  pages={3077-3085}
}

@inproceedings{NEURIPS2024_12d3e63b,
 author = {Peng, Yingzhe and Hao, Chenduo and Hu, Xinting and Peng, Jiawei and Geng, Xin and Yang, Xu},
 booktitle = {Advances in Neural Information Processing Systems},
 doi = {10.52202/079017-0312},
 editor = {A. Globerson and L. Mackey and D. Belgrave and A. Fan and U. Paquet and J. Tomczak and C. Zhang},
 pages = {9773--9800},
 publisher = {Curran Associates, Inc.},
 title = {LIVE: Learnable In-Context Vector for Visual Question Answering},
 url = {https://proceedings.neurips.cc/paper_files/paper/2024/file/12d3e63be5574088f7c1bbc9162060bf-Paper-Conference.pdf},
 volume = {37},
 year = {2024}
}

@misc{li2025catpcontextuallyadaptivetoken,
      title={CATP: Contextually Adaptive Token Pruning for Efficient and Enhanced Multimodal In-Context Learning}, 
      author={Yanshu Li and Jianjiang Yang and Zhennan Shen and Ligong Han and Haoyan Xu and Ruixiang Tang},
      year={2025},
      eprint={2508.07871},
      archivePrefix={arXiv},
      primaryClass={cs.CV},
      url={https://arxiv.org/abs/2508.07871}, 
}

@article{li2025cama,
  title={CAMA: Enhancing Multimodal In-Context Learning with Context-Aware Modulated Attention},
  author={Li, Yanshu and Yang, JianJiang and Li, Bozheng and Tang, Ruixiang},
  journal={arXiv preprint arXiv:2505.17097},
  year={2025}
}

@article{gupta2025meta,
  title={Meta-Adaptive Prompt Distillation for Few-Shot Visual Question Answering},
  author={Gupta, Akash and Storkey, Amos and Lapata, Mirella},
  journal={arXiv preprint arXiv:2506.06905},
  year={2025}
}

@inproceedings{NEURIPS2024_a0303731,
 author = {Lauren\c{c}on, Hugo and Tronchon, L\'{e}o and Cord, Matthieu and Sanh, Victor},
 booktitle = {Advances in Neural Information Processing Systems},
 doi = {10.52202/079017-2789},
 editor = {A. Globerson and L. Mackey and D. Belgrave and A. Fan and U. Paquet and J. Tomczak and C. Zhang},
 pages = {87874--87907},
 publisher = {Curran Associates, Inc.},
 title = {What matters when building vision-language models?},
 url = {https://proceedings.neurips.cc/paper_files/paper/2024/file/a03037317560b8c5f2fb4b6466d4c439-Paper-Conference.pdf},
 volume = {37},
 year = {2024}
}

@inproceedings{NEURIPS2020_1457c0d6,
 author = {Brown, Tom and Mann, Benjamin and Ryder, Nick and Subbiah, Melanie and Kaplan, Jared D and Dhariwal, Prafulla and Neelakantan, Arvind and Shyam, Pranav and Sastry, Girish and Askell, Amanda and Agarwal, Sandhini and Herbert-Voss, Ariel and Krueger, Gretchen and Henighan, Tom and Child, Rewon and Ramesh, Aditya and Ziegler, Daniel and Wu, Jeffrey and Winter, Clemens and Hesse, Chris and Chen, Mark and Sigler, Eric and Litwin, Mateusz and Gray, Scott and Chess, Benjamin and Clark, Jack and Berner, Christopher and McCandlish, Sam and Radford, Alec and Sutskever, Ilya and Amodei, Dario},
 booktitle = {Advances in Neural Information Processing Systems},
 editor = {H. Larochelle and M. Ranzato and R. Hadsell and M.F. Balcan and H. Lin},
 pages = {1877--1901},
 publisher = {Curran Associates, Inc.},
 title = {Language Models are Few-Shot Learners},
 url = {https://proceedings.neurips.cc/paper_files/paper/2020/file/1457c0d6bfcb4967418bfb8ac142f64a-Paper.pdf},
 volume = {33},
 year = {2020}
}

@article{jiang2024many,
  title={Many-shot in-context learning in multimodal foundation models},
  author={Jiang, Yixing and Irvin, Jeremy and Wang, Ji Hun and Chaudhry, Muhammad Ahmed and Chen, Jonathan H and Ng, Andrew Y},
  journal={arXiv preprint arXiv:2405.09798},
  year={2024}
}

@article{huang2025mimicking,
  title={Mimicking or Reasoning: Rethinking Multi-Modal In-Context Learning in Vision-Language Models},
  author={Huang, Chengyue and Zhu, Yuchen and Zhu, Sichen and Xiao, Jingyun and Andrade, Moises and Chopra, Shivang and Kira, Zsolt},
  journal={arXiv preprint arXiv:2506.07936},
  year={2025}
}

@inproceedings{NEURIPS2024_deeb4d6b,
 author = {Qin, Libo and Chen, Qiguang and Fei, Hao and Chen, Zhi and Li, Min and Che, Wanxiang},
 booktitle = {Advances in Neural Information Processing Systems},
 doi = {10.52202/079017-3916},
 editor = {A. Globerson and L. Mackey and D. Belgrave and A. Fan and U. Paquet and J. Tomczak and C. Zhang},
 pages = {123207--123236},
 publisher = {Curran Associates, Inc.},
 title = {What Factors Affect Multi-Modal In-Context Learning? An In-Depth Exploration},
 url = {https://proceedings.neurips.cc/paper_files/paper/2024/file/deeb4d6bdb5860fd7faf321dd5486d25-Paper-Conference.pdf},
 volume = {37},
 year = {2024}
}

@inproceedings{chen2025can,
  title={Can Multimodal Large Language Models Truly Perform Multimodal In-Context Learning?},
  author={Chen, Shuo and Han, Zhen and He, Bailan and Liu, Jianzhe and Buckley, Mark and Qin, Yao and Torr, Philip and Tresp, Volker and Gu, Jindong},
  booktitle={2025 IEEE/CVF Winter Conference on Applications of Computer Vision (WACV)},
  pages={6000--6010},
  year={2025},
  organization={IEEE}
}

@inproceedings{doveh2024towards,
  title={Towards multimodal in-context learning for vision and language models},
  author={Doveh, Sivan and Perek, Shaked and Mirza, M Jehanzeb and Lin, Wei and Alfassy, Amit and Arbelle, Assaf and Ullman, Shimon and Karlinsky, Leonid},
  booktitle={European Conference on Computer Vision},
  pages={250--267},
  year={2024},
  organization={Springer}
}

@inproceedings{chen2024internvl,
  title={Internvl: Scaling up vision foundation models and aligning for generic visual-linguistic tasks},
  author={Chen, Zhe and Wu, Jiannan and Wang, Wenhai and Su, Weijie and Chen, Guo and Xing, Sen and Zhong, Muyan and Zhang, Qinglong and Zhu, Xizhou and Lu, Lewei and others},
  booktitle={Proceedings of the IEEE/CVF conference on computer vision and pattern recognition},
  pages={24185--24198},
  year={2024}
}

@inproceedings{dong2024survey,
  title={A survey on in-context learning},
  author={Dong, Qingxiu and Li, Lei and Dai, Damai and Zheng, Ce and Ma, Jingyuan and Li, Rui and Xia, Heming and Xu, Jingjing and Wu, Zhiyong and Chang, Baobao and others},
  booktitle={Proceedings of the 2024 conference on empirical methods in natural language processing},
  pages={1107--1128},
  year={2024}
}

@inproceedings{duan2024vlmevalkit,
  title={Vlmevalkit: An open-source toolkit for evaluating large multi-modality models},
  author={Duan, Haodong and Yang, Junming and Qiao, Yuxuan and Fang, Xinyu and Chen, Lin and Liu, Yuan and Dong, Xiaoyi and Zang, Yuhang and Zhang, Pan and Wang, Jiaqi and others},
  booktitle={Proceedings of the 32nd ACM international conference on multimedia},
  pages={11198--11201},
  year={2024}
}

@misc{zheng2025easyr1,
  title={EasyR1: An Efficient, Scalable, Multi-Modality RL Training Framework},
  author={Yaowei Zheng and Junting Lu and Shenzhi Wang and Zhangchi Feng and Dongdong Kuang and Yuwen Xiong and Richong Zhang},
  howpublished={\url{https://github.com/hiyouga/EasyR1}},
  year={2025}
}

@article{sheng2024hybridflow,
  title   = {HybridFlow: A Flexible and Efficient RLHF Framework},
  author  = {Guangming Sheng and Chi Zhang and Zilingfeng Ye and Xibin Wu and Wang Zhang and Ru Zhang and Yanghua Peng and Haibin Lin and Chuan Wu},
  year    = {2024},
  journal = {arXiv preprint arXiv: 2409.19256}
}

@inproceedings{zheng2024llamafactory,
  title={Llamafactory: Unified efficient fine-tuning of 100+ language models},
  author={Zheng, Yaowei and Zhang, Richong and Zhang, Junhao and Ye, Yanhan and Luo, Zheyan},
  booktitle={Proceedings of the 62nd annual meeting of the association for computational linguistics (volume 3: system demonstrations)},
  pages={400--410},
  year={2024}
}

@article{wang2025context,
  title={IN-CONTEXT LEARNING AS GENERAL-PURPOSE LEARNING: A COMPREHENSIVE SURVEY AND NEW PERSPECTIVES},
  author={Wang, Fan and Yu, Bo and Shao, Pengtao and Chen, Zhiyuan and Chou, Hungchyun and Kang, Liuwang and Shao, Shenbao and Xia, Xuan and Ding, Ning and Xiong, Haoyi and others},
  journal={Authorea Preprints},
  year={2025},
  publisher={Authorea}
}

@article{kuhnle2017shapeworld,
  title={Shapeworld-a new test methodology for multimodal language understanding},
  author={Kuhnle, Alexander and Copestake, Ann},
  journal={arXiv preprint arXiv:1704.04517},
  year={2017}
}


\appendix

\section{Detailed Data Construction}

\subsection{Process Typical VQA Datasets}
\label{suppl:process-typical-VQA}

For typical VQA datasets, underlying rules are derived from existing annotations. We group samples with identical question texts and treat each unique question as an underlying rule that defines the visual–answer mapping. To ensure sufficient coverage and stability, we prioritize frequently occurring questions and manually filter them based on whether their semantics are suitable for rule abstraction. For samples sharing the same question but having different answers, we remove low-frequency answers to keep the answer distribution under each rule roughly balanced.

During MICL episode construction, the original question is replaced with a generic instruction to avoid revealing task-specific information. Following prior MICL settings~\cite{zong2025vl}, we further reduce textual leakage by remapping non-numerical answers to nonsensical tokens, while retaining original answers for mathematical questions where numerical values are essential.

In addition, we randomly select some rule groups to construct paired-image matching episodes. Each sample contains two images and a binary answer indicating whether the two images produce the same answer under the original underlying rule, with \emph{Yes} for matching answers and \emph{No} otherwise.

\subsection{Rule Designs for Synthetic Data}
\label{suppl:rule-designs-for-synthetic-data}

For Super-CLEVR-3D, we extend the CLEVR-style rule design used in the VL-ICL benchmark to a more challenging setting. Each rule group is defined by a randomly selected object attribute and value, denoted as $(\mathrm{attr}, \mathrm{value})$. For each sample, the image is drawn from Super-CLEVR-3D, while the question specifies an additional attribute–value pair $(\mathrm{attr}_q, \mathrm{value}_q)$. The answer is the number of objects in the image whose $\mathrm{attr}$ equals $\mathrm{value}$ and whose $\mathrm{attr}_q$ equals $\mathrm{value}_q$. This construction introduces a hidden attribute–value constraint shared across the episode, requiring the model to jointly induce the underlying rule and integrate sample-specific query information for reasoning.
Due to the sparsity caused by attribute-based filtering in Super-CLEVR-3D, raw sampling often results in answers concentrated at small values (e.g., 0 or 1). To avoid biased answer distributions that may weaken numerical reasoning, we apply additional filtering and sampling strategies to balance answer frequencies within each rule group.

For mathematical reasoning, we manually design two types of synthetic datasets. The first contains images with a single number, where each rule group is defined by a simple arithmetic mapping applied to the number, such as addition, subtraction, or multiplication within a small range. The second contains images depicting a numerical sequence, randomly sampled as either arithmetic or geometric progressions. The image shows the first three terms, and the answer corresponds to the next term in the sequence. In both cases, the underlying mathematical relation defines the rule shared by all samples within a group.

\section{All Prompt Templates}
\label{suppl:all-prompt-templates}

\subsection{Unified Prompt Template for Case II MICL}

\begin{framed}
\noindent\textbf{Direct-Answer Mode}

\noindent You are a visual reasoning assistant. Below are several examples. Each example contains one or more images, a question, and an answer. Across all examples, there is a hidden rule or pattern that links the images (possibly in combination with the question) to the correct answer. \\
\{ \textit{task description} \} \\
Your task is to study the examples, infer the underlying rule that maps from images and questions to answers, and then apply this rule to a new input consisting of image(s) and a question. \\
\#\#\# Examples: \\
...... \\
\\
\#\#\#\# Example \{ \textit{index} \}: \\
Images: \{ \textit{images of support index} \} \\
Question: \{ \textit{question of support index} \} \\
Answer: \{ \textit{answer of support index} \} \\
\\
...... \\
\#\#\# Now, here is a new input: \\
Images: \{ \textit{query images} \} \\
Question: \{ \textit{query question} \} \\
Answer:
\end{framed}

\begin{framed}
\noindent\textbf{Think-Then-Answer Mode}

\noindent You are a visual reasoning assistant. Below are several examples. Each example contains one or more images, a question, and an answer. Across all examples, there is a hidden rule or pattern that links the images (possibly in combination with the question) to the correct answer. \\
\{ \textit{task description} \} \\
Your task is to study the examples, infer the underlying rule that maps from images and questions to answers, and then apply this rule to a new input consisting of image(s) and a question. \\
When answering the new input, first think step by step inside \texttt{<think>...</think>} tags, and then output only the final answer inside \texttt{<answer>...</answer>} tags. \\
\#\#\# Examples: \\
...... \\
\\
\#\#\#\# Example \{ \textit{index} \}: \\
Images: \{ \textit{images of support index} \} \\
Question: \{ \textit{question of support index} \} \\
Answer: \{ \textit{answer of support index} \} \\
\\
...... \\
\#\#\# Now, here is a new input: \\
Images: \{ \textit{query images} \} \\
Question: \{ \textit{query question} \}
\end{framed}

\subsection{Prompt for Caption Generation}

\begin{framed}
\noindent You are a multimodal captioning assistant. \\
Generate precise, detailed, and natural-language captions for all kinds of images  (real photos, charts, math figures, or synthetic scenes). \\
\\
\#\#\# Core Rules \\
1. **Faithfulness** — Describe only what is visible or explicitly given. No guessing or opinions. \\
2. **Structured Detail** — Begin with an overall scene summary, then note key objects, attributes, relations, and visible text. \\
3. **Annotation Coverage** — The caption must include *all factual details* from the annotation JSON (objects, values, labels, text, etc.) naturally in the description. Never mention the annotation itself. \\
4. **Formatting** — \\
   - Transcribe visible text exactly. \\
   - Use LaTeX for math (e.g., `$E = mc^2$`). \\
   - For tables/plots, describe structure, axes, legends, and trends. \\
   - For 3D/synthetic scenes, note materials, shapes, colors, and spatial layout. \\
5. **Style** — Neutral, factual, and concise. Output only the caption text. \\
\\
Below is an image and its corresponding annotation. \\
Please generate a comprehensive, faithful, and natural-language caption following the system rules above. \\
\#\#\#\# Image: \\
\{ \textit{image} \} \\
\#\#\#\# Annotation (JSON): \\
```json \\
\{ \textit{annotation} \}
\end{framed}

\subsection{Prompt for Caption Rule-Preserving Judgment}

\begin{framed}
\noindent You are an impartial evaluator. You will receive: \\
1. A caption describing an image. \\
2. A JSON annotation containing key factual and semantic details of that image. \\
Determine whether the caption sufficiently covers the annotation: \\
- If all annotation details are explicitly stated in the caption, output "Yes". \\
- If the caption does not state certain details verbatim but they can be reasonably inferred from its meaning, also output "Yes". \\
- If the caption includes more information than the annotation, still output "Yes". \\
- Output "No" only when some annotation information cannot be found or reasonably inferred from the caption, or is contradicted. \\
Respond with only one word: "Yes" or "No". \\
\\
Caption: \{ \textit{caption} \} \\
Annotation (JSON): \{ \textit{annotation} \} \\
Output:
\end{framed}

\subsection{Prompt for Rule-Induction Generation}

\begin{framed}
\noindent You are a reasoning model designed to carefully work through problems step by step before providing a final answer. \\
\\
Task: Below are several examples. Each example contains one or more images, a question, and an answer. Across all examples, there is a hidden rule or pattern that links the images (possibly in combination with the question) to the correct answer. \\
Your task is to study the examples, infer the underlying rule that maps from images and questions to answers, and then apply this rule to a new input consisting of image(s) and a question. \\
\\
For every image you receive a caption. Treat each caption as a faithful, vivid visual rendering - act as if you are directly perceiving the scene. Do NOT say or refer to phrases such as "based on the caption", "the caption says", "I read the description", or "visualizing the image"; instead write observations as if you are seeing the image ("The image shows...", "There are three red objects..."). \\
\\
Output requirements (strict - produce nothing else): \\
1. Please reason step by step inside \texttt{<think>...</think>}. \\
2. After \texttt{<think>}, output only the final answer inside \texttt{<answer>...</answer>}. \\
3. NEVER output any extraneous commentary, data, or formatting beyond these tags. \\
\\
\#\#\# Examples: \\
...... \\
\#\#\#\# Example \{ \textit{index} \} \\
Visual Contents: \\
- Image Captions: \{ \textit{images captions of support index} \} \\
Question:\{ \textit{question of support index} \} \\
Answer:\{ \textit{answer of support index} \} \\
...... \\
\#\#\# Now, here is a new input: \\
Visual Contents: \\
- Image Captions: \{ \textit{query images captions} \} \\
Question:\{ \textit{query question} \}
\end{framed}

\subsection{Prompt for Rule Consistency Evaluation}

\begin{framed}
\noindent\textbf{Prompt for Rule Consistency Evaluation}

\noindent You are an evaluator whose job is to determine whether a provided Chain-of-Thought (CoT)  \\
follows a specified rule. \\
\#\#\# Task \\
You will be given: \\
1. A natural-language rule description. \\
2. A description of how this rule is applied in the current task setting. \\
3. A CoT produced by a model. \\
Your job is to check whether the CoT correctly applies the rule under this specific setting. \\
\#\#\# Output \\
Output only: \\
- "Yes" if the CoT fully aligns with the rule and the described setting. \\
- "No" otherwise. \\
Do not provide explanations or any additional text. \\
\\
Here is the rule and the current task setting. \\
Rule \\
\{ \textit{rule description} \} \\
Task Setting \\
\{ \textit{task setting text} \} \\
CoT to Evaluate \\
\{ \textit{CoT} \} \\
Does the CoT correctly follow the rule under this setting? \\
Output only "Yes" or "No".
\end{framed}

The \textit{task setting text} in the rule consistency evaluation prompt is treated as a variable component, instantiated according to the specific configuration of each MICL episode. 

We first determine whether the episode corresponds to a paired-image matching setting, where each sample contains two images and the answer is a binary label indicating whether the two images yield the same outcome under the original underlying rule. This condition is indicated by the variable \textit{is pair}. We then consider whether the episode applies nonsensical token remapping to replace original answers, which is indicated by the variable \textit{use nonsense}.

The \textit{task setting text} is instantiated differently under each combination of these variables, ensuring that the rule consistency evaluation prompt accurately reflects the episode structure while avoiding unintended information leakage.


\begin{framed}
\noindent\textbf{Task Setting Text}

\noindent \{\% \textbf{if \textit{is pair}} \%\} \\
In this task, each context sample contains a pair of images. \\
Under the base rule, each image would independently receive an answer. \\
For paired-image tasks, the final answer is determined by comparing the two answers: \\
If the two answers are identical, the correct result is 'Yes'. \\
If they differ, the correct result is 'No'. \\
The CoT must follow this paired-image rule. \\
\{\% \textbf{elif \textit{use nonsense}} \%\} \\
In this task, answers that follow the base rule are replaced by meaningless tokens \\
using a predefined mapping. The CoT must apply the base rule correctly, \\
but must express all answers using the nonsense words instead of meaningful ones. \\
\{\% \textbf{else} \%\} \\
In this task, answers follow the base rule directly with no modification. \\
\{\% \textbf{endif} \%\}
\end{framed}

\subsection{Prompt for Rewriting}

\begin{framed}
\noindent You will receive a snippet of text that references a "description" or "caption" of an image. Your task is to produce a **nearly identical** version of that text with **minimal** changes, focusing on the following: \\
\\
1. **Replace references to "description", "caption" and "rationale"** with wording that references **"the image."** \\
   - For example, "The description says..." could become "The image shows..." \\
   - "The caption suggests..." could become "The image suggests..." \\
   - "Based on the rationale..." could become "Based on the image..." \\
   - Make sure the replacement sounds natural but does **not** otherwise change the meaning. \\
\\
2. **Preserve all line breaks, punctuation, and spacing** as much as possible, and make **no additional edits** outside of these replacements.   \\
\\
3. You should only output the rewritten content. \\
\\
Here is the input: \{ \textit{CoT} \}"
\end{framed}

\section{Experiment Details}

\subsection{Detailed Data Split for SFT and DAPO}
\label{suppl:detailed-data-split}

In our two-stage fine-tuning strategy, we separate the data used for supervised fine-tuning and reinforcement learning to avoid overlap and to maintain a balanced split. This design prevents the RL stage from being overly constrained by supervised signals and preserves sufficient room for policy exploration. All training data are obtained from the Explicit Rule-Induction Supervision generation process applied to rule-centric MICL episodes. Owing to differences in data sources and rule design strategies, subsets exhibit substantial variation in the success rate of generating valid rule-induction chains. To maintain a globally reasonable SFT–DAPO data ratio while fully utilizing the generated supervision, we adopt an adaptive data split across subsets. Specifically, subsets with higher induction-generation success rates allocate a smaller proportion of data to SFT and reserve more episodes for DAPO training. In contrast, subsets with lower success rates allocate a larger fraction of data to SFT. 

\subsection{Implementation Details}
\label{suppl:implementation-details}

All inference and evaluation for our model and baselines are conducted using VLMEvalKit~\cite{duan2024vlmevalkit}. We extend the framework by integrating the VL-ICL and TrueMICL benchmarks, enabling unified multimodal inference and evaluation. Inference-related settings are summarized in Table~\ref{tab:inference-hparams}.

Supervised fine-tuning is performed with LLaMA-Factory~\cite{zheng2024llamafactory}, and the corresponding hyperparameters are reported in Table~\ref{tab:sft-hparams}. For the reinforcement learning stage, we adopt EasyR1~\cite{zheng2025easyr1,sheng2024hybridflow} to conduct multimodal DAPO training, with hyperparameters listed in Table~\ref{tab:dapo-hparams}. All experiments are carried out on 4 NVIDIA RTX A6000 GPUs.

\begin{table}
\centering
\begin{tabular}{l|c}
\toprule
Hyperparameter & Value \\
\midrule
temperature & 0.0 \\
min p & 0.0 \\
top p & 1.0 \\
top k & -1 \\
repetition penalty & 1.0 \\
frequency penalty & 0.0 \\
presence penalty & 0.0 \\
max token & 2048 \\
\bottomrule
\end{tabular}
\caption{Inference hyperparameters.}
\label{tab:inference-hparams}
\end{table}

\begin{table}
\centering
\begin{tabular}{l|c}
\toprule
Hyperparameter & Value \\
\midrule
batch size & 128 \\
learning rate & 1e-5 \\
train epochs & 6 \\
lr scheduler type & cosine \\
warmup ratio & 0.1 \\
training precision & bfloat16 \\
image min pixels & 3136 \\
image max pixels & 313600 \\
freeze vision tower & true \\
freeze multi modal projector & true \\
\bottomrule
\end{tabular}
\caption{Supervised fine-tuning hyperparameters.}
\label{tab:sft-hparams}
\end{table}

\begin{table}
\centering
\begin{tabular}{l|c}
\toprule
Hyperparameter & Value \\
\midrule
max prompt length & 3072 \\
max response length & 2048 \\
rollout batch size & 64 \\
global batch size & 16 \\
min pixels & 3136 \\
max pixels & 313600 \\
kl coef & 1e-3 \\
training precision & bfloat16 \\
rollout n & 8 \\
rollout temperature & 1.0 \\
rollout top p & 1.0 \\
total epochs & 4 \\
filter low (DAPO) & 0.01 \\
filter high (DAPO) & 0.99 \\
learning rate & 1e-6 \\
weight decay & 1e-2 \\
lr warmup ratio & 0.0 \\
\bottomrule
\end{tabular}
\caption{DAPO hyperparameters.}
\label{tab:dapo-hparams}
\end{table}

\subsection{Complete Main Results}
\label{suppl:complete-main-results}

The complete main experimental results are reported in Table~\ref{tab:main-detailed}.

\begin{table*}[t]
\centering
{
\small
\setlength{\tabcolsep}{5pt}
\begin{tabular}{ll|ccccc|c||ccccc|c}
\toprule

\multicolumn{1}{c}{\multirow{3}{*}{Method}} & \multicolumn{1}{c}{\multirow{3}{*}{Mode}} &
\multicolumn{6}{c}{Open MI} & \multicolumn{6}{c}{CLEVR} \\
\cmidrule(lr){3-8} \cmidrule(lr){9-14}
\multicolumn{2}{c}{} &
\multicolumn{1}{c}{0} & \multicolumn{1}{c}{1} & \multicolumn{1}{c}{2} &
\multicolumn{1}{c}{4} & \multicolumn{1}{c}{5} & \multicolumn{1}{c}{Avg} &
\multicolumn{1}{c}{2} & \multicolumn{1}{c}{3} & \multicolumn{1}{c}{4} &
\multicolumn{1}{c}{5} & \multicolumn{1}{c}{6} & \multicolumn{1}{c}{Avg} \\
\midrule
OpenFlamingo-9B & \textcolor{basecolor}{\textit{answer}} & 0.00 & 5.17 & 49.67 & 80.17 & 80.50 & 43.10 & 24.00 & 24.67 & 24.17 & 26.67 & 24.83 & 24.87 \\
IDEFICS3-8B-Llama3 & \textcolor{basecolor}{\textit{answer}} & 0.00 & 0.00 & 6.83 & 33.00 & 39.33 & 15.83 & 57.00 & 54.00 & 54.50 & 57.83 & 60.83 & 56.83 \\
Phi-3.5-Vision & \textcolor{basecolor}{\textit{answer}} & 0.00 & 2.50 & 20.33 & 29.33 & 29.83 & 16.40 & 32.17 & 26.83 & 28.33 & 26.00 & 28.33 & 28.33 \\
LLaVA-OneVision-7B & \textcolor{basecolor}{\textit{answer}} & 0.00 & 23.83 & 78.00 & 85.50 & 87.33 & 54.93 & 46.33 & 48.83 & 51.17 & 50.50 & 49.83 & 49.33 \\
OpenVLThinker-3B & \textit{reasoning} & 0.00 & 4.17 & 27.17 & 30.00 & 30.67 & 18.40 & 56.33 & 57.17 & 53.00 & 55.83 & 53.33 & 55.13 \\
VLAA-Thinker-3B & \textit{reasoning} & 0.00 & 4.50 & 24.50 & 35.00 & 40.67 & 20.93 & 56.83 & 56.67 & 54.67 & 51.50 & 48.67 & 53.67 \\
\rowcolor{highlightcolor} Qwen2.5-VL-3B & \textcolor{basecolor}{\textit{answer}} & 0.00 & 12.17 & 81.50 & 90.33 & 89.50 & 54.70 & 51.00 & 52.67 & 53.33 & 54.67 & 55.17 & 53.37 \\
ERIM-3B (Ours) & \textbf{\textit{induction}} & 0.00 & \textbf{83.67} & \textbf{96.67} & \textbf{97.67} & \textbf{96.17} & \textbf{74.83} & \textbf{64.00} & \textbf{64.33} & \textbf{65.33} & \textbf{66.83} & \textbf{63.83} & \textbf{64.87} \\
\midrule

\multicolumn{1}{c}{\multirow{3}{*}{Method}} & \multicolumn{1}{c}{\multirow{3}{*}{Mode}} &
\multicolumn{6}{c}{Operator Induction} & \multicolumn{6}{c}{Interleaved Operator Induction} \\
\cmidrule(lr){3-8} \cmidrule(lr){9-14}
\multicolumn{2}{c}{} &
\multicolumn{1}{c}{2} & \multicolumn{1}{c}{3} & \multicolumn{1}{c}{4} &
\multicolumn{1}{c}{5} & \multicolumn{1}{c}{6} & \multicolumn{1}{c}{Avg} &
\multicolumn{1}{c}{2} & \multicolumn{1}{c}{3} & \multicolumn{1}{c}{4} &
\multicolumn{1}{c}{5} & \multicolumn{1}{c}{6} & \multicolumn{1}{c}{Avg} \\
\midrule
OpenFlamingo-9B & \textcolor{basecolor}{\textit{answer}} & 8.33 & 7.78 & 6.11 & 5.00 & 8.33 & 7.11 & 7.22 & 5.56 & 2.78 & 3.33 & 6.11 & 5.00 \\
IDEFICS3-8B-Llama3 & \textcolor{basecolor}{\textit{answer}} & 0.00 & 0.00 & 1.11 & 1.67 & 1.67 & 0.89 & 5.00 & 4.44 & 7.22 & 7.22 & 10.56 & 6.89 \\
Phi-3.5-Vision & \textcolor{basecolor}{\textit{answer}} & 10.56 & 15.56 & 11.11 & 6.11 & 7.78 & 10.22 & 8.89 & 7.22 & 8.33 & 6.11 & 5.56 & 7.22 \\
LLaVA-OneVision-7B & \textcolor{basecolor}{\textit{answer}} & 43.89 & 50.56 & 54.44 & 49.44 & 55.00 & 50.67 & 40.00 & 47.78 & 47.22 & 46.67 & 46.11 & 45.56 \\
OpenVLThinker-3B & \textit{reasoning} & 73.33 & 82.22 & 80.00 & 79.44 & 82.22 & 79.44 & 60.56 & 57.78 & 56.67 & 69.44 & 66.67 & 62.22 \\
VLAA-Thinker-3B & \textit{reasoning} & 67.78 & 76.11 & 83.33 & 83.89 & 80.56 & 78.33 & 60.00 & 61.67 & 68.33 & 69.44 & 75.00 & 66.89 \\
\rowcolor{highlightcolor} Qwen2.5-VL-3B & \textcolor{basecolor}{\textit{answer}} & 67.78 & 74.44 & 74.44 & 80.56 & 76.67 & 74.78 & 54.44 & 60.56 & 62.22 & 58.89 & 66.11 & 60.44 \\
ERIM-3B (Ours) & \textbf{\textit{induction}} & \textbf{88.89} & \textbf{92.22} & \textbf{90.56} & \textbf{96.67} & \textbf{97.22} & \textbf{93.11} & \textbf{84.44} & \textbf{82.78} & \textbf{87.22} & \textbf{79.44} & \textbf{83.89} & \textbf{83.56} \\
\midrule

\multicolumn{1}{c}{\multirow{3}{*}{Method}} & \multicolumn{1}{c}{\multirow{3}{*}{Mode}} &
\multicolumn{6}{c}{Matching MI} & \multicolumn{6}{c}{Clock} \\
\cmidrule(lr){3-8} \cmidrule(lr){9-14}
\multicolumn{2}{c}{} &
\multicolumn{1}{c}{0} & \multicolumn{1}{c}{1} & \multicolumn{1}{c}{2} &
\multicolumn{1}{c}{4} & \multicolumn{1}{c}{5} & \multicolumn{1}{c}{Avg} &
\multicolumn{1}{c}{2} & \multicolumn{1}{c}{3} & \multicolumn{1}{c}{4} &
\multicolumn{1}{c}{5} & \multicolumn{1}{c}{6} & \multicolumn{1}{c}{Avg} \\
\midrule
OpenFlamingo-9B & \textcolor{basecolor}{\textit{answer}} & 32.08 & 50.00 & 50.00 & 50.00 & 50.00 & 46.42 & 5.00 & 7.00 & 3.00 & 3.00 & 5.00 & 4.60 \\
IDEFICS3-8B-Llama3 & \textcolor{basecolor}{\textit{answer}} & 62.58 & 50.17 & 50.08 & 49.75 & 50.17 & 52.55 & 3.00 & 3.00 & 5.00 & 8.00 & 6.00 & 5.00 \\
Phi-3.5-Vision & \textcolor{basecolor}{\textit{answer}} & 68.42 & 49.75 & 49.92 & 50.08 & 50.08 & 53.65 & 7.00 & 5.00 & 11.00 & 6.00 & 6.00 & 7.00 \\
LLaVA-OneVision-7B & \textcolor{basecolor}{\textit{answer}} & 84.00 & 55.83 & 56.67 & 56.75 & 54.58 & 61.57 & 21.00 & 14.00 & 18.00 & 16.00 & 14.00 & 16.60 \\
OpenVLThinker-3B & \textit{reasoning} & \textbf{77.58} & 56.75 & 55.83 & 58.00 & 56.83 & 61.00 & 44.00 & 61.00 & 74.00 & 71.00 & \textbf{74.00} & 64.80 \\
VLAA-Thinker-3B & \textit{reasoning} & 75.50 & 60.67 & 62.50 & 63.08 & 65.33 & 65.42 & 42.00 & 52.00 & 52.00 & 60.00 & 54.00 & 52.00 \\
\rowcolor{highlightcolor} Qwen2.5-VL-3B & \textcolor{basecolor}{\textit{answer}} & 67.67 & 52.08 & 52.17 & 53.42 & 54.50 & 55.97 & 32.00 & 41.00 & 45.00 & 51.00 & 49.00 & 43.60 \\
ERIM-3B (Ours) & \textbf{\textit{induction}} & 64.17 & \textbf{76.25} & \textbf{77.08} & \textbf{81.58} & \textbf{82.08} & \textbf{76.23} & \textbf{78.00} & \textbf{71.00} & \textbf{83.00} & \textbf{77.00} & 71.00 & \textbf{76.00} \\
\midrule

\multicolumn{1}{c}{\multirow{3}{*}{Method}} & \multicolumn{1}{c}{\multirow{3}{*}{Mode}} &
\multicolumn{6}{c}{Palindrome} & \multicolumn{6}{c}{CLEVR (TrueMICL)} \\
\cmidrule(lr){3-8} \cmidrule(lr){9-14}
\multicolumn{2}{c}{} &
\multicolumn{1}{c}{2} & \multicolumn{1}{c}{3} & \multicolumn{1}{c}{4} &
\multicolumn{1}{c}{5} & \multicolumn{1}{c}{6} & \multicolumn{1}{c}{Avg} &
\multicolumn{1}{c}{2} & \multicolumn{1}{c}{3} & \multicolumn{1}{c}{4} &
\multicolumn{1}{c}{5} & \multicolumn{1}{c}{6} & \multicolumn{1}{c}{Avg} \\
\midrule
OpenFlamingo-9B & \textcolor{basecolor}{\textit{answer}} & 6.00 & 9.00 & 9.00 & 11.00 & 11.00 & 9.20 & 30.00 & 34.00 & 34.00 & 33.00 & 30.00 & 32.20 \\
IDEFICS3-8B-Llama3 & \textcolor{basecolor}{\textit{answer}} & 35.00 & 45.00 & 40.00 & 39.00 & 39.00 & 39.60 & \textbf{35.00} & 31.00 & 39.00 & 31.00 & \textbf{39.00} & 35.00 \\
Phi-3.5-Vision & \textcolor{basecolor}{\textit{answer}} & 11.00 & 6.00 & 15.00 & 8.00 & 14.00 & 10.80 & 25.00 & 29.00 & 29.00 & 35.00 & 30.00 & 29.60 \\
LLaVA-OneVision-7B & \textcolor{basecolor}{\textit{answer}} & \textbf{87.00} & \textbf{87.00} & \textbf{91.00} & \textbf{92.00} & \textbf{90.00} & \textbf{89.40} & 30.00 & 32.00 & 38.00 & 35.00 & 37.00 & 34.40 \\
OpenVLThinker-3B & \textit{reasoning} & 71.00 & 79.00 & 78.00 & 75.00 & 77.00 & 76.00 & 16.00 & 18.00 & 15.00 & 16.00 & 18.00 & 16.60 \\
VLAA-Thinker-3B & \textit{reasoning} & 39.00 & 44.00 & 35.00 & 46.00 & 48.00 & 42.40 & 11.00 & 12.00 & 6.00 & 12.00 & 12.00 & 10.60 \\
\rowcolor{highlightcolor} Qwen2.5-VL-3B & \textcolor{basecolor}{\textit{answer}} & 41.00 & 35.00 & 30.00 & 28.00 & 27.00 & 32.20 & 12.00 & 19.00 & 24.00 & 27.00 & 31.00 & 22.60 \\
ERIM-3B (Ours) & \textbf{\textit{induction}} & 55.00 & 68.00 & 73.00 & 81.00 & 77.00 & 70.80 & \textbf{35.00} & \textbf{37.00} & \textbf{46.00} & \textbf{44.00} & 32.00 & \textbf{38.80} \\

\bottomrule
\end{tabular}
}
\caption{Complete main experimental results comparing ERIM with baseline methods across all evaluation benchmarks.}
\label{tab:main-detailed}
\end{table*}

\subsection{Complete Training Strategy Ablation Results}
\label{suppl:complete-ablation-strategy-results}

The complete training strategy ablation results are reported in Table~\ref{tab:ablation-strategy-detailed}.

\begin{table*}[t]
\centering
{
\small
\setlength{\tabcolsep}{5pt}
\begin{tabular}{lcc|ccccc|c||ccccc|c}
\toprule
\multicolumn{1}{c}{\multirow{2}{*}{Method}} & \multicolumn{2}{c}{Strategy} &
\multicolumn{6}{c}{Open MI} & \multicolumn{6}{c}{CLEVR} \\
\cmidrule(lr){2-3} \cmidrule(lr){4-9} \cmidrule(lr){10-15}
\multicolumn{1}{c}{} &
\multicolumn{1}{c}{SFT} & \multicolumn{1}{c}{RL} & 
\multicolumn{1}{c}{0} & \multicolumn{1}{c}{1} & \multicolumn{1}{c}{2} &
\multicolumn{1}{c}{4} & \multicolumn{1}{c}{5} & \multicolumn{1}{c}{Avg} &
\multicolumn{1}{c}{2} & \multicolumn{1}{c}{3} & \multicolumn{1}{c}{4} &
\multicolumn{1}{c}{5} & \multicolumn{1}{c}{6} & \multicolumn{1}{c}{Avg} \\
\midrule
Qwen2.5VL & \ding{55} & \ding{55} & 0.00 & 12.17 & 81.50 & 90.33 & 89.50 & 54.70 & 51.00 & 52.67 & 53.33 & 54.67 & 55.17 & 53.37 \\
+Direct-SFT & Direct & \ding{55} & 0.00 & \textbf{98.50} & \textbf{99.67} & \textbf{100.00} & \textbf{99.67} & \textbf{79.57} & 60.17 & 61.17 & 61.17 & 62.83 & 59.83 & 61.03 \\
+Think-RL & \ding{55} & \ding{51} & 0.00 & 91.83 & 98.83 & 99.33 & 99.33 & 77.87 & 49.00 & 51.17 & 51.33 & 50.67 & 51.00 & 50.63 \\
+ERIM & Mixed & \ding{51} & 0.00 & 83.67 & 96.67 & 97.67 & 96.17 & 74.83 & \textbf{64.00} & \textbf{64.33} & \textbf{65.33} & \textbf{66.83} & \textbf{63.83} & \textbf{64.87} \\
\midrule
\multicolumn{1}{c}{\multirow{2}{*}{Method}} & \multicolumn{2}{c}{Strategy} &
\multicolumn{6}{c}{Operator Induction} & \multicolumn{6}{c}{Interleaved Operator Induction} \\
\cmidrule(lr){2-3} \cmidrule(lr){4-9} \cmidrule(lr){10-15}
\multicolumn{1}{c}{} &
\multicolumn{1}{c}{SFT} & \multicolumn{1}{c}{RL} & 
\multicolumn{1}{c}{2} & \multicolumn{1}{c}{3} & \multicolumn{1}{c}{4} &
\multicolumn{1}{c}{5} & \multicolumn{1}{c}{6} & \multicolumn{1}{c}{Avg} &
\multicolumn{1}{c}{2} & \multicolumn{1}{c}{3} & \multicolumn{1}{c}{4} &
\multicolumn{1}{c}{5} & \multicolumn{1}{c}{6} & \multicolumn{1}{c}{Avg} \\
\midrule
Qwen2.5VL & \ding{55} & \ding{55} & 67.78 & 74.44 & 74.44 & 80.56 & 76.67 & 74.78 & 54.44 & 60.56 & 62.22 & 58.89 & 66.11 & 60.44 \\
+Direct-SFT & Direct & \ding{55} & 62.22 & 74.44 & 77.22 & 81.67 & 79.44 & 75.00 & 67.22 & 73.33 & 75.56 & 76.11 & 78.89 & 74.22 \\
+Think-RL & \ding{55} & \ding{51} & 76.11 & 86.67 & \textbf{90.56} & 93.33 & 91.11 & 87.56 & 68.89 & 73.89 & 76.67 & \textbf{81.11} & \textbf{84.44} & 77.00 \\
+ERIM & Mixed & \ding{51} & \textbf{88.89} & \textbf{92.22} & \textbf{90.56} & \textbf{96.67} & \textbf{97.22} & \textbf{93.11} & \textbf{84.44} & \textbf{82.78} & \textbf{87.22} & 79.44 & 83.89 & \textbf{83.56} \\
\midrule
\multicolumn{1}{c}{\multirow{2}{*}{Method}} & \multicolumn{2}{c}{Strategy} &
\multicolumn{6}{c}{Matching MI} & \multicolumn{6}{c}{Clock} \\
\cmidrule(lr){2-3} \cmidrule(lr){4-9} \cmidrule(lr){10-15}
\multicolumn{1}{c}{} &
\multicolumn{1}{c}{SFT} & \multicolumn{1}{c}{RL} & 
\multicolumn{1}{c}{0} & \multicolumn{1}{c}{1} & \multicolumn{1}{c}{2} &
\multicolumn{1}{c}{4} & \multicolumn{1}{c}{5} & \multicolumn{1}{c}{Avg} &
\multicolumn{1}{c}{2} & \multicolumn{1}{c}{3} & \multicolumn{1}{c}{4} &
\multicolumn{1}{c}{5} & \multicolumn{1}{c}{6} & \multicolumn{1}{c}{Avg} \\
\midrule
Qwen2.5VL & \ding{55} & \ding{55} & 67.67 & 52.08 & 52.17 & 53.42 & 54.50 & 55.97 & 32.00 & 41.00 & 45.00 & 51.00 & 49.00 & 43.60 \\
+Direct-SFT & Direct & \ding{55} & 80.75 & 85.83 & 86.67 & 85.25 & 85.50 & 84.80 & 31.00 & 44.00 & 37.00 & 46.00 & 43.00 & 40.20 \\
+Think-RL & \ding{55} & \ding{51} & \textbf{90.00} & \textbf{89.75} & \textbf{89.67} & \textbf{90.00} & \textbf{89.92} & \textbf{89.87} & 64.00 & \textbf{72.00} & 81.00 & \textbf{87.00} & \textbf{83.00} & \textbf{77.40} \\
+ERIM & Mixed & \ding{51} & 64.17 & 76.25 & 77.08 & 81.58 & 82.08 & 76.23 & \textbf{78.00} & 71.00 & \textbf{83.00} & 77.00 & 71.00 & 76.00 \\
\midrule
\multicolumn{1}{c}{\multirow{2}{*}{Method}} & \multicolumn{2}{c}{Strategy} &
\multicolumn{6}{c}{Palindrome} & \multicolumn{6}{c}{CLEVR (TrueMICL)} \\
\cmidrule(lr){2-3} \cmidrule(lr){4-9} \cmidrule(lr){10-15}
\multicolumn{1}{c}{} &
\multicolumn{1}{c}{SFT} & \multicolumn{1}{c}{RL} & 
\multicolumn{1}{c}{2} & \multicolumn{1}{c}{3} & \multicolumn{1}{c}{4} &
\multicolumn{1}{c}{5} & \multicolumn{1}{c}{6} & \multicolumn{1}{c}{Avg} &
\multicolumn{1}{c}{2} & \multicolumn{1}{c}{3} & \multicolumn{1}{c}{4} &
\multicolumn{1}{c}{5} & \multicolumn{1}{c}{6} & \multicolumn{1}{c}{Avg} \\
\midrule
Qwen2.5VL & \ding{55} & \ding{55} & 41.00 & 35.00 & 30.00 & 28.00 & 27.00 & 32.20 & 12.00 & 19.00 & 24.00 & 27.00 & 31.00 & 22.60 \\
+Direct-SFT & Direct & \ding{55} & 39.00 & 34.00 & 37.00 & 33.00 & 32.00 & 35.00 & 31.00 & 31.00 & 31.00 & 30.00 & 31.00 & 30.80 \\
+Think-RL & \ding{55} & \ding{51} & \textbf{63.00} & 66.00 & 68.00 & 57.00 & 62.00 & 63.20 & 23.00 & 24.00 & 22.00 & 19.00 & 21.00 & 21.80 \\
+ERIM & Mixed & \ding{51} & 55.00 & \textbf{68.00} & \textbf{73.00} & \textbf{81.00} & \textbf{77.00} & \textbf{70.80} & \textbf{35.00} & \textbf{37.00} & \textbf{46.00} & \textbf{44.00} & \textbf{32.00} & \textbf{38.80} \\
\bottomrule
\end{tabular}
}
\caption{Complete ablation results analyzing the impact of different training strategies.}
\label{tab:ablation-strategy-detailed}
\end{table*}

\begin{table*}[t]
\centering
\begin{tabular}{ll|ccccc|c}
\toprule
\multicolumn{1}{c}{\multirow{3}{*}{Method}} & \multicolumn{1}{c}{\multirow{3}{*}{Mode}} & \multicolumn{6}{c}{ShapeWorld} \\
\cmidrule(lr){3-8}
\multicolumn{2}{c}{} &
\multicolumn{1}{c}{2} & \multicolumn{1}{c}{3} & \multicolumn{1}{c}{4} &
\multicolumn{1}{c}{5} & \multicolumn{1}{c}{6} & \multicolumn{1}{c}{Avg} \\
\midrule
OpenFlamingo-9B & \textcolor{basecolor}{\textit{answer}} & 36.67  & 23.33  & 21.11  & 35.56  & 30.00  & 29.33 \\
Phi-3.5-Vision & \textcolor{basecolor}{\textit{answer}} & 41.11  & 25.56  & 26.67  & 28.89  & 30.00  & 30.45 \\
OpenVLThinker-3B & \textit{reasoning} & 37.78  & 26.67  & 44.44  & 37.78  & 40.00  & 37.33 \\
VLAA-Thinker-3B & \textit{reasoning} & 40.00  & 33.33  & \textbf{46.67}  & 42.22  & 36.67  & 39.78 \\
\rowcolor{highlightcolor} Qwen2.5-VL-3B & \textcolor{basecolor}{\textit{answer}} & 30.00  & 33.33  & 35.56  & 33.33  & 31.11  & 32.67 \\
ERIM-3B (Ours) & \textbf{\textit{induction}} & \textbf{41.11}  & \textbf{34.44}  & 42.22  & \textbf{46.67}  & \textbf{42.22}  & \textbf{41.33} \\
\midrule
\bottomrule
\end{tabular}
\caption{Results on out-of-domain evaluation benchmark. We report accuracy (\%) of different methods under varying n-shot settings, along with the average performance over n-shots.}
\label{tab:out-of-domain}
\end{table*}


\section{Out-of-Domain Evaluation}

To further evaluate whether our method generalizes beyond the data distributions seen during training and in-domain evaluation, we construct an additional \emph{out-of-domain} Case II MICL benchmark from a multimodal dataset that does not appear in either our training data or the main evaluation benchmarks. 
Following the construction principle of existing Case II benchmarks, we manually organize this dataset into rule-governed MICL episodes and evaluate several major baselines together with our final model, \textbf{ERIM-3B}. 
This setting provides a stricter test of whether the model can transfer its learned rule-induction behavior to a new multimodal distribution, rather than benefiting from familiarity with previously seen datasets or benchmark formats.

\subsection{Data Construction}

For out-of-domain evaluation, we use \textbf{ShapeWorld}~\cite{kuhnle2017shapeworld}, a software-rendered dataset of 2D scenes with controllable data distributions and detailed annotations. 
Its clean visual structure and precise attribute annotations make it suitable for constructing a new Case II MICL benchmark with well-defined underlying rules. 
Following the same general design principle as the CLEVR subset in VL-ICL~\cite{zong2025vl}, we define the underlying rule in each episode as the mapping from the attribute-value relation described in the question to the number of objects in the image that satisfy the corresponding condition. 
In this construction, we focus on two attributes, namely \emph{shape} and \emph{color}, and organize samples into support-query episodes according to the shared underlying rule. 
As a result, the final ShapeWorld benchmark serves as an out-of-domain Case II evaluation set that differs from both our training data and the benchmark distributions used in the main experiments.

\subsection{Result Analysis}

Table~\ref{tab:out-of-domain} reports the out-of-domain evaluation results on the constructed ShapeWorld benchmark. 
ERIM-3B achieves the best overall performance, obtaining the highest average accuracy among all compared methods. 
Although some baselines remain competitive under individual shot settings, our method delivers the strongest overall results and achieves the best accuracy in most settings. 
These results show that the advantage of ERIM is not limited to the training distribution or the in-domain benchmarks considered in the main paper. 
Instead, the model retains strong performance when transferred to a previously unseen multimodal dataset with a different visual domain and rule instantiation, demonstrating better generalization of explicit rule induction. 
This out-of-domain evaluation therefore provides additional evidence that our method learns more transferable rule-induction behavior for Case II MICL, rather than merely adapting to the specific distributions of the training and evaluation benchmarks.

\end{document}